%% file: main.tex
\documentclass[nohyperref]{article}

\usepackage{microtype}
\usepackage{graphicx}
\usepackage{subfigure}
\usepackage{booktabs} 
\usepackage{amsfonts,amssymb}

\usepackage{hyperref}

\usepackage[accepted]{icml2022}

\usepackage{amsmath}
\usepackage{amssymb}
\usepackage{mathtools}
\usepackage{amsthm}

\usepackage{algorithm}
\usepackage{algorithmic}

\usepackage[english]{babel}
\usepackage[utf8]{inputenc} 
\usepackage[T1]{fontenc}    
\usepackage{hyperref}       
\usepackage{booktabs}       
\usepackage{amsfonts}       
\usepackage{nicefrac}       
\usepackage{microtype}      
\usepackage{xcolor}         

\usepackage{amsmath}
\usepackage{algorithmic}
\usepackage{algorithm}
\usepackage{array}
\usepackage{float}
\usepackage{multirow}
\usepackage{epsfig}
\usepackage{epstopdf}
\usepackage{amssymb}
\usepackage{tabularx}
\usepackage{lineno}
\usepackage{longtable}
\usepackage{booktabs}
\usepackage{bm}
\usepackage{rotating}
\usepackage{mathrsfs}
\usepackage{color}
\usepackage{amsthm}
\usepackage{graphicx}
\usepackage{pythonhighlight}
\usepackage{bm}

\usepackage[capitalize,noabbrev]{cleveref}

\theoremstyle{plain}

\theoremstyle{definition}

\theoremstyle{remark}

\usepackage[textsize=tiny]{todonotes}

\icmltitlerunning{\textbf{MetAug: Contrastive Learning via Meta Feature Augmentation}}

\begin{document}

\twocolumn[
\icmltitle{MetAug: Contrastive Learning via Meta Feature Augmentation}



\icmlsetsymbol{equal}{*}

\begin{icmlauthorlist}
\icmlauthor{Jiangmeng Li}{equal,yyy,comp,compp}
\icmlauthor{Wenwen Qiang}{equal,yyy,comp,compp}
\icmlauthor{Changwen Zheng}{yyy,compp}
\icmlauthor{Bing Su}{sch,bjk}
\icmlauthor{Hui Xiong}{yyyy,yyyyy}
\end{icmlauthorlist}
\icmlaffiliation{yyy}{Science \& Technology on Integrated Information System Laboratory, Institute of Software Chinese Academy of Sciences, Beijing, China}
\icmlaffiliation{comp}{University of Chinese Academy of Sciences, Beijing, China}
\icmlaffiliation{compp}{Southern Marine Science and Engineering Guangdong Laboratory (Guangzhou), Guangdong, China.}
\icmlaffiliation{sch}{Gaoling School of Artificial Intelligence,  Renmin University of China, Beijing, China}
\icmlaffiliation{bjk}{Beijing Key Laboratory of Big Data Management and Analysis Methods, Beijing, China}
\icmlaffiliation{yyyy}{Thrust of Artificial Intelligence, The Hong Kong University of Science and Technology (Guangzhou), Guangzhou, China}
\icmlaffiliation{yyyyy}{Department of Computer Science & Engineering, The Hong Kong University of Science and Technology, Hong Kong SAR, China}
\icmlcorrespondingauthor{Bing Su}{subingats@gmail.com}

\icmlkeywords{Contrastive Learning, Self-Supervised Learning, Classification, Feature Augmentation, Meta Learning}

\vskip 0.3in
]


\newcommand\blfootnote[1]{%
	\begingroup
	\renewcommand\thefootnote{}\footnote{#1}%
	\addtocounter{footnote}{-1}%
	\endgroup
}


\blfootnote{*Equal contribution $^1$ Science \& Technology on Integrated Information System Laboratory, Institute of Software Chinese Academy of Sciences, Beijing, China $^2$ University of Chinese Academy of Sciences, Beijing, China $^3$ Southern Marine Science and Engineering Guangdong Laboratory (Guangzhou), Guangdong, China. $^4$ Gaoling School of Artificial Intelligence, Renmin University of China, Beijing, China $^5$ Beijing Key Laboratory of Big Data Management and Analysis Methods, Beijing, China $^6$ Thrust of Artificial Intelligence, The Hong Kong University of Science and Technology (Guangzhou), Guangzhou, China $^7$ Department of Computer Science Engineering, The Hong Kong University of Science and Technology, Hong Kong SAR, China. Correspondence to: Bing Su <subingats@gmail.com>. \\\\Proceedings of the 39 th International Conference on Machine Learning, Baltimore, Maryland, USA, PMLR 162, 2022. Copyright 2022 by the author(s).}

\input{abstract}
\input{introduction}

\input{related_works}
\input{method}
\input{experiments}
\input{conclusion}



\bibliography{mainbib}
\bibliographystyle{icml2022}

\input{appendix}

\end{document}

%% file: abstract.tex
\begin{abstract}
What matters for contrastive learning? We argue that contrastive learning heavily relies on \textit{informative} features, or ``hard'' (positive or negative) features. Early works include more informative features by applying complex data augmentations and large batch size or memory bank, and recent works design elaborate sampling approaches to explore informative features. The key challenge toward exploring such features is that the source multi-view data is generated by applying random data augmentations, making it infeasible to always add useful information in the augmented data. Consequently, the informativeness of features learned from such augmented data is limited. In response, we propose to directly augment the features in latent space, thereby learning discriminative representations without a large amount of input data. We perform a meta learning technique to build the augmentation generator that updates its network parameters by considering the performance of the encoder. However, insufficient input data may lead the encoder to learn collapsed features and therefore malfunction the augmentation generator. A new margin-injected regularization is further added in the objective function to avoid the encoder learning a degenerate mapping. To contrast all features in one gradient back-propagation step, we adopt the proposed optimization-driven unified contrastive loss instead of the conventional contrastive loss. Empirically, our method achieves state-of-the-art results on several benchmark datasets.

\end{abstract}

%% file: introduction.tex
\section{Introduction}\label{sec:introduction}
Contrastive learning methods have achieved empirical success in computer vision \cite{2005Chopra, 2006Hadsell}. Under the setting of self-supervised learning (SSL), recent researches demonstrate the superiority of contrastive methods \cite{hjelm2018learning, Tian2019Contrastive, 2020Debiased, 2020Hard}. Typically, these approaches learn features by contrasting different views (e.g., different random data augmentations) of the image in hidden space. We recap the preliminaries of the conventional contrastive learning paradigm: every two views of the \textit{same} image are considered to be a positive pair, and every two views of the \textit{different} images are considered to be a negative pair; the contrastive loss \cite{2010Michael, 2018RepresentationOord} guides the learned features to bring \textit{positive} pairs together and push \textit{negative} pairs farther apart.

However, this learning paradigm suffers from the need for a large number of pairs to contrast, e.g., large batch size or memory bank size, because many pairs are not informative to the model, i.e., positive pairs are pretty close and negative pairs are already very apart in hidden space. These pairs have few contributions to the optimization. Contrastive methods need numerous pairs and expect to collect informative ones, and therefore complex data augmentations (e.g., jittering, random cropping, separating color channels, etc.) \cite{2019Philip, chen2020simple} and large-scale memory banks \cite{Tian2019Contrastive, 2020Kaiming} are effective in improving the performance of contrastive models on downstream tasks.

The success of the recent works depends on the elaborate selection of informative negative pairs \cite{2020Debiased, 2020Hard}. These methods focus on designing sampling strategies to assign larger weights to informative pairs, which rely on enough and informative positive pairs and do not need large amounts of negative pairs. When the number of pairs to contrast is limited, the contrastive loss may cause conventional contrastive learning approaches to learn collapsed features \cite{2021Barlow, 2020Bootstrap}, e.g., outputting the same feature vector for all images. 

Nowadays, many researchers have noticed potential environmental problems brought by training deep learning models \cite{2021AGDLJingjing}, for instance, \cite{2019EnergyStrubell} reports a remarkable example that the carbon dioxide emissions generated by training a Transformer \cite{2017Ashish} is equivalent to 200 round trips between San Francisco and New York by plane. Therefore, we motivate our method to perform an efficient self-supervised contrastive approach to learn anti-collapse and discriminative features based on a restricted amount of images in a training epoch (e.g., small batch size) and plain neural networks with limited parameters. Much research effort has been devoted to strong augmentations on \textit{data}, but the informativeness of the features learned from the augmented data is hard to exactly measure, since the data is fed into mapping-agnostic deep neural networks to generate the features. Instead, we directly tackle augmentations on features and show that appropriate feature augmentations can sharply improve the optimization.

To this end, we propose \textit{\textbf{Met}a Feature \textbf{Aug}mentation} (MetAug), which learns view-specific encoders (with projection heads) and auxiliary meta feature augmentation generators (MAGs) by margin-injected meta feature augmentation and optimization-driven unified contrast. Suppose the input data has $\bm{M}$ views, and the multi-view data is fed into the encoder to generate the latent features. We initialize $\bm{M}$ neural networks as MAGs for views, which are used to augment the features of each view. We contrast all original and augmented features for bi-optimization training. Through such a learning paradigm, MetAug can improve the performance of self-supervised contrastive learning.


To learn anti-collapse and discriminative features from a restricted amount of images, MetAug relies on two key ingredients: 1) margin-injected meta feature augmentation, where MAGs use the performance of the encoder in one iteration to improve the view-specific feature augmentations for the next iteration. In this way, MAGs promote the encoder to efficiently explore the discriminative information of the input. For the original features and the augmented features generated by MAGs, we inject a margin $\mathcal{R_\sigma}$ between the similarities of them, which avoids the instance-level feature collapse; 2) optimization-driven unified contrast, which contrasts all features in one gradient back-propagation step. Such proposed contrast can also amplify the impact of the instance similarity that deviates far from the optimum and weaken the impact of the instance similarity that is close to the optimum. We conduct head-to-head comparisons on various benchmark datasets, which prove the effectiveness of margin-injected meta feature augmentation and optimization-driven unified contrast. \textbf{Contributions}:

\begin{itemize}
	\item We propose margin-injected meta feature augmentation, which directly augments the latent features to generate informative and anti-collapse features. Benefiting from such features, encoders can efficiently capture discriminative information.
	\item We propose optimization-driven unified contrast to include all available features in one step of back-propagation and weight the similarities of paired features by measuring their contributions to optimization.
	\item Empirically, MetAug improves the downstream task performance on different benchmark datasets.
\end{itemize}

%% file: related_works.tex
\section{Related works}
{\bf{Self-supervised learning.}} Under the setting of unsupervised learning, SSL methods have achieved impressive success, which constructs auxiliary tasks to learn discriminative information from the unlabeled inputs. Deep InfoMax \cite{hjelm2018learning} explores to maximize the mutual information between an input and the output of a deep neural network encoder by different mutual information estimations. CPC \cite{2018RepresentationOord} proposes to adopt noise-contrastive estimation (NCE) \cite{2010Michael} as the contrastive loss to train the model to measure the mutual information of multiple views deduced by the Kullback-Leibler divergence \cite{2003Goldbberger}. CMC \cite{Tian2019Contrastive} and AMDIM \cite{2019Philip} employ contrastive learning on the multi-view data. SwAV \cite{2020Mathilde} compares the cluster assignments under different views instead of directly comparing features by using more views (e.g., six views). SimCLR \cite{chen2020simple} and MoCo \cite{2020Kaiming} use large batch or memory bank to enlarge the amount of available negative features to learn good representations. Instead of exploring informative features by adopting various data augmentations and enlarging the number of features, our method focuses on straightforwardly generating informative features to contrast.

\begin{figure*}
	\centering
	\includegraphics[width=0.7\textwidth]{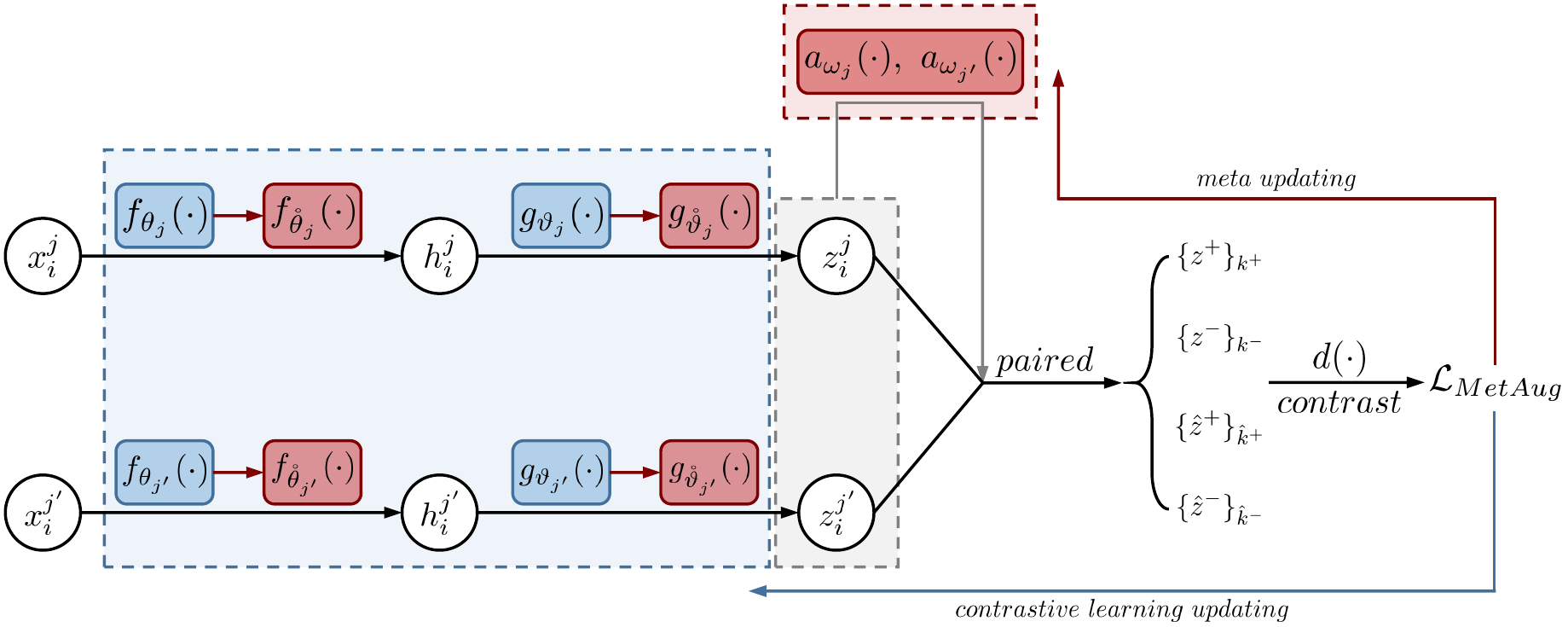}
	\vskip -0.1in
	\caption{MetAug's architecture. \textit{Dashed blue} box represents the data encoding process, and \textit{dashed red} box represents the meta feature augmentation. In training, we first fix $a_{\omega_j}$ and $a_{\omega_{j^\prime}}$, and then train $f_{\theta_j}(\cdot)$, $g_{\vartheta_j}(\cdot)$, $f_{\theta_{j^\prime}}(\cdot)$ and $g_{\vartheta_{j^\prime}}(\cdot)$ by using $\mathcal{L}_{MetAug}$. Next, we fix the encoders and projection heads, and train $a_{\omega_j}$ and $a_{\omega_{j^\prime}}$ in a meta updating manner. The networks are iteratively trained until convergence.}
	\label{fig:algoframe}
	\vspace{-0.5cm}
\end{figure*}

Recent works explore imposing stronger constraints on the conventional contrastive learning paradigm or propose alternative loss functions (instead of contrastive loss). DebiasedCL\cite{2020Debiased} and HardCL \cite{2020Hard} consider to directly collect informative features to contrast by designing sampling strategies, which are inspired by positive-unlabeled learning methods \cite{2008Charles, 2014Marthinus}. Motivated by \cite{2008Sridharan}, \cite{2020Tsai} proposes an information theoretical framework for SSL, which, guided by the theory, uses information bottleneck to restrict the learned features and maintain the sufficient self-supervision. BYOL \cite{2020Bootstrap}, W-MSE \cite{2020WhiteningErmolov}, and Barlow Twins \cite{2021Barlow} present a crucial issue that insufficient self-supervision (e.g., not enough negative features) may lead to the feature collapse in hidden space. To tackle the mentioned issue, we propose a new margin-injected regularization in meta feature augmentation to avoid generating degenerate features. DACL \cite{2021Vikas} proposes a new data augmentation that applies to domain-agnostic problems. LooC \cite{2021Tete} learns to capture varying and invariant factors for visual representations by constructing separate embedding spaces for each augmentation. These methods explore informative features from the perspective of data augmentation, while our straightforward idea behind our method is to augment features in the latent space.

{\bf{Meta learning.}} The objective of meta learning is to automatically learn the \textit{learning algorithm}. Early works \cite{Succ2002On, 2002Bengio, 2014Schmidhuber} aim to guide the model (e.g., neural network) to learn prior knowledge about \textit{how to learn new knowledge}, so that the model can efficiently learn new knowledge, e.g., the model can be quickly fine-tuned to specific downstream tasks with few training steps and achieve good performance. Recently, researchers explored to use meta learning to find optimal hyper-parameters \cite{2017MetaLi} and appropriately initialize a neural network for few-shot learning \cite{2017ModelFinn, 2017Jake, 2016Oriol}. Recent approaches \cite{2016LearningChen, 2016DecoupledJaderberg, 2018VisualMa, 2019SelfLiu} have focused on learning optimizers or generating a gradient-driven loss for deep neural networks in the field of NLP, computer vision, etc.

%% file: method.tex
\section{Method}
\label{sec:method}
Our goal is to learn representations that capture information shared between multiple different views by performing self-supervised contrastive learning. Formally, we suppose the input multi-view dataset as ${X} = \left\{{X_1, X_2, ..., X_N}\right\}$, where $N$ denotes the number of samples. $X_i$ represents a collection of $M$ views of the sample, where $i \in \left\{ {1, ..., N} \right\}$. For each sample $X_i$, we denote $x_i$ as a random variable representing views following $x_i \thicksim \mathcal{P}(X_i)$, and ${x_i^j}$ denotes the $j$-th view of the $i$-th sample, where $j \in \left\{ {1, ..., M} \right\}$.

\begin{figure*}
	\centering
	\includegraphics[width=0.85\textwidth]{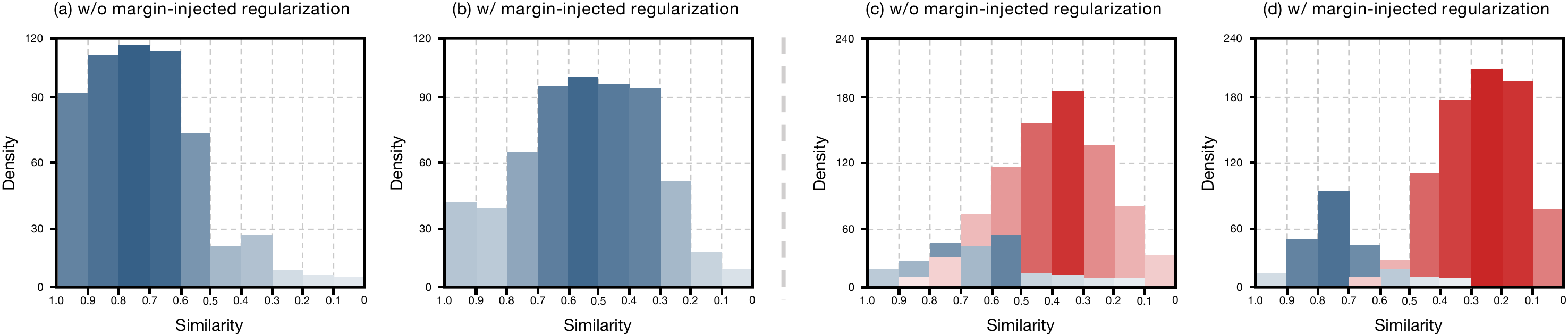}
	\vskip -0.1in
	\caption{Similarity histograms obtained by our method (with or without margin-injected regularization) on CIFAR-10. (a) and (b) demonstrate the summarized similarity of positives (i.e., $\{\hat{z}^+\}$) that include original features and augmented features. (c) and (d) demonstrate the statistical results of the original features learned by our model. \textit{Blue} histograms represent the similarity between the features of the same image's views, and \textit{red} histograms represent the similarity between the features of the different images' views.}
	\label{fig:augsimstats}
	\vspace{-0.5cm}
\end{figure*}

\subsection{Contrastive learning preliminary}
We recap the preliminaries of contrastive learning\cite{Tian2019Contrastive, chen2020simple}: the foundational idea behind contrastive learning is to learn an embedding that maximizes agreement between the views of the same sample and separates the views of different samples in latent space. Given a multi-view dataset $X$, we treat pairs of the views of the same sample $\{x_i^j, x_i^{j'}\}$, where $j, j' \in \{1, ..., M\}$, as \textit{positives}, versus pairs of the views of the different samples $\{x_i^j, x_{i'}^{j'}\}$, where $i \neq i'$, as \textit{negatives}. To impose contrastive learning, we feed the input $x_i^j$ into a view-specific encoder $f_{\theta_j}(\cdot)$ to learn a representation $h_i^j$, and $h_i^j$ is mapped into a feature $z_i^j$ by a projection head $g_{\vartheta_j}(\cdot)$, where $\theta_j$ and $\vartheta_j$ are the network parameters of $f_{\theta_j}(\cdot)$ and $g_{\vartheta_j}(\cdot)$, respectively. A discriminating function $d(\cdot)$ is adopted to measure the similarity of $\{z_i^j, z_{i'}^{j'}\}$, where $i \neq i'$. The encoder $f_{\theta_j}(\cdot)$ and projection head $g_{\vartheta_j}(\cdot)$ are trained by using a contrastive loss \cite{2018RepresentationOord}, which is formulated as follows:
\begin{equation}
		{\mathcal{L}} = -\mathop{\mathbb{E}}_{X_S} {\left[\log \frac{{d\left(\{z^+\}\right)}}{{ d\left(\{z^+\}\right) + \sum\limits_{k = 1}^K {{{d}\left(\{z^-\}_k\right)} } }}\right]}
		\label{eq:cl}
\end{equation}
where $X_S = \{\{z^+\}, \{z^-\}_1, \{z^-\}_2, ..., , \{z^-\}_k\}$ is a set of \textit{pairs} randomly sampled from $X$, which includes a positive $\{z^+\}$ and $K$ negatives $\{z^-\}_k$, $k \in \{1, ..., K\}$, because contrastive loss can only use \textit{one} positive in an iteration. In test, the projection head $g_{\vartheta_j}(\cdot)$ is discarded, and the representation $h_i^j$ is directly used for downstream tasks.

\subsection{Margin-injected meta feature augmentation}
Recent contrastive methods rely on complex data augmentations to increase the informativeness of views. Yet this lack of guidance approach leads to the demand for a large number of training data (e.g., large batch size and memory bank). We propose a meta feature augmentation method, which creates informative augmented features by updating parameters of its own network according to the performance (gradient) of the encoder (see Appendix \ref{sec:rethinkaug} for our rethinking of augmented features). A visualization of the overall MetAug architecture is shown in Figure \ref{fig:algoframe}.

To this end, we build a group of MAGs $a_{\omega}(\cdot) = \{a_{\omega_1}(\cdot), ..., a_{\omega_M}(\cdot)\}$ for all $M$ views, where $\omega = \{\omega_1, ..., \omega_M\}$. To be simplified, we define $f_\theta$ and $g_\vartheta$ as the groups of view-specific encoders and projection heads, respectively, i.e., $\theta = \{\theta_1, ..., \theta_M\}$ and $\vartheta = \{\vartheta_1, ..., \vartheta_M\}$.

In training, the encoders $f_{\theta}(\cdot)$ and the projection heads $g_{\vartheta}(\cdot)$ are trained alongside the MAG $a_{\omega}(\cdot)$ (with the network parameters $\omega$). Following the protocol of meta learning \cite{2017ModelFinn, 2019SelfLiu}, we firstly train $f_{\theta}(\cdot)$ and $g_{\vartheta}(\cdot)$ under the learning paradigm of self-supervised contrastive learning. Then, $a_{\omega}(\cdot)$ is updated by computing its gradients with respect to the performance of $f_{\theta}(\cdot)$ and $g_{\vartheta}(\cdot)$. Here, we measure the performance of $f_{\theta}(\cdot)$ and $g_{\vartheta}(\cdot)$ by the gradients of them when the corresponding contrastive loss is back-propagated. Concretely, all of $f_{\theta}(\cdot)$, $g_{\vartheta}(\cdot)$, and $a_{\omega}(\cdot)$ are iteratively trained until convergence.

Specifically, we first update network parameters $\theta$ and $\vartheta$ of the encoders and projection heads by adopting the conventional contrastive loss. Then, we train the MAG $a_{\omega}(\cdot)$ in a meta learning manner. We encourage the augmented features to be informative, and the encoders $f_{\theta}(\cdot)$ can better explore the discriminative information by jointly using the original and augmented features to contrast. Hence, the performance of the encoders would be promoted on the same training data. To update network parameters $\omega$ of the $a_{\omega}(\cdot)$, we formalize the meta updating objective as follows:
\begin{equation}
	\mathop{\arg\min}_{\omega} \mathcal{L}\bigg(\Big\{g_{\mathring{\vartheta}}\big(f_{\mathring{\theta}}\big( \widetilde{X} \big)\big), a_\omega\Big(g_{\mathring{\vartheta}}\big(f_{\mathring{\theta}}\big( \widetilde{X} \big)\big)\Big) \Big\}\bigg)
	\label{eq:metaomegaj}
\end{equation}
where $\widetilde{X}$ represents a minibatch sampled from the training dataset $X$, $\Big\{g_{\mathring{\vartheta}}\big(f_{\mathring{\theta}}\big( \widetilde{X} \big)\big), a_\omega\Big(g_{\mathring{\vartheta}}\big(f_{\mathring{\theta}}\big( \widetilde{X} \big)\big)\Big) \Big\}$ denotes a set including both original features and meta augmented features. $\mathring{\vartheta}$ and $\mathring{\theta}$ represent the parameter sets of the encoders and projection heads, respectively, which are computed by the updating of one gradient back-propagation:
\begin{equation}
	\begin{aligned}
		&\mathring{\theta} = \theta - \ell\cdot\nabla_{\theta} \mathcal{L}\bigg(\Big\{g_{\vartheta}\big(f_{\theta}\big( \widetilde{X} \big)\big), a_\omega\Big(g_{\vartheta}\big(f_{\theta}\big( \widetilde{X} \big)\big)\Big) \Big\}\bigg) \\
		&\mathring{\vartheta} = \vartheta - \ell\cdot\nabla_{\vartheta} \mathcal{L}\bigg(\Big\{g_{\vartheta}\big(f_{\theta}\big( \widetilde{X} \big)\big), a_\omega\Big(g_{\vartheta}\big(f_{\theta}\big( \widetilde{X} \big)\big)\Big) \Big\}\bigg) 
	\end{aligned}
	\label{eq:fastweight}
\end{equation}
where $\ell$ is the learning rate shared between $\theta$ and $\vartheta$. The idea behind the meta updating objective is that we perform the second-derivative technique \cite{2017ModelFinn, 2018FineYabin, 2019SelfLiu} to train $a_{\omega}(\cdot)$. Specifically, a derivative over the derivative (Hessian matrix) of the combination $\{\theta, \vartheta\}$ is used to update $\omega$, where $\{\theta, \vartheta\}$ is a parameter set conjoining $\theta$ and $\vartheta$. We compute the derivative with respect to $\omega$ by using a retained computational graph of $\{\theta, \vartheta\}$.

However, in practice, we find a critical issue: when the original features are not informative enough, large gradients are difficult to generate by contrasting the uninformative features, the MAGs $a_{\omega}(\cdot)$ are inclined to create collapsed augmented features, e.g., the augmented features and the original features are very similar. We consider the reason for the feature collapse is that small gradient changes of the encoders alongside projection heads $g_\vartheta(f_\theta(\cdot))$ lead to the update step-size of $a_{\omega}(\cdot)$ to become extensively small, which leaves the optimization of $a_{\omega}(\cdot)$ to fall into a local optimum. The augmented features are such that without any extra useful information. To tackle this issue, we further inject a margin to encourage $a_{\omega}(\cdot)$ to generate more complex and informative augmented features, which can be considered as a regularization term in the meta updating objective. See Figure \ref{fig:augsimstats}(a) for the details of the augmented feature collapse issue, and we observe that, without margin-injected regularization, MAGs tend to generate collapsed features that are very similar with the original features. Formally, we formulate the approach to generate margins for $a_{\omega}(\cdot)$ by
\begin{equation}
	\begin{array}{lr}
		\sigma^+ = \min\Big[\min\big(\big\{d(\{z^+\}_{k^+})\big\}\big), \max\big(\big\{d(\{z^-\}_{k^-})\big\}\big)\Big] &  \\ \\
		\sigma^- = \max\Big[\min\big(\big\{d(\{z^+\}_{k^+})\big\}\big), \max\big(\big\{d(\{z^-\}_{k^-})\big\}\big)\Big] &  
	\end{array}
	\label{eq:marginapp}
\end{equation}
where $\big\{d(\{z^+\}_{k^+})\big\}$ is a set of the outputs (similarities) of positives computed by the discriminating function $d(\cdot)$, and $k^+ \in \{1, ..., K^+\}$ where $K^+$ represents the number of positives in a minibatch. $\Big\{d(\{z^-\}_{k^-})\Big\}$ is a set of the discriminating outputs of negatives, and $k^- \in \{1, ..., K^-\}$ where $K^-$ represents the number of negatives. Note that only original features are used in Equation \ref{eq:marginapp}. We call the formulated margin generation approach as \textit{"Large"}, and we also propose two more approaches, called \textit{"Medium"} and \textit{"Small"}. In Appendix \ref{sec:marginvariant}, we conduct comparisons to evaluate the effects of the three margin generation approaches. We inject the margins between the augmented features and original features by adding a regularization term in the meta updating objective, and the regularization is defined as:
\begin{equation}
	\begin{aligned}
		\mathcal{R_\sigma} = &\frac{1}{\hat{K}^+}\sum_{\hat{k}^+ = 1}^{\hat{K}^+}\Big[ d(\{\hat{z}^+\}_{\hat{k}^+}) - \sigma^+ \Big]_+\\
		&+ \frac{1}{\hat{K}^-}\sum_{\hat{k}^- = 1}^{\hat{K}^-}\Big[ \sigma^- - d(\{\hat{z}^-\}_{\hat{k}^-}) \Big]_+
	\end{aligned}
	\label{eq:marginreg}
\end{equation}
where $\{\hat{z}^+\}_{\hat{k}^+}$ denotes a positive that includes one \textit{original} feature and one \textit{augmented} feature, and $\hat{K}^+$ denotes the number of such positives. $\{\hat{z}^-\}_{\hat{k}^-}$ represents likewise one of $\hat{K}^-$ negatives, each of which includes one \textit{original} feature and one \textit{augmented} feature. $[\cdot]_{+}$ denotes the cut-off-at-zero function, which is defined as $[a]_{+} = \max(a, 0)$. We then integrate such regularization to the updating of $\omega$ by
\begin{equation}
	\begin{aligned}
		\omega \leftarrow \omega - \ell^\prime\cdot\nabla_{\omega} \mathcal{L}\bigg(\Big\{g_{\mathring{\vartheta}}\big(f_{\mathring{\theta}}\big( \widetilde{X} \big)\big), \\ a_\omega\Big(g_{\mathring{\vartheta}}\big(f_{\mathring{\theta}}\big( \widetilde{X} \big)\big)\Big) \Big\}\bigg) + \alpha \cdot \mathcal{R_\sigma}
	\end{aligned}
	\label{eq:fastweightomega}
\end{equation}
where $\ell^\prime$ represents the learning rate of $\omega$, and $\alpha$ is a hyperparameter balancing the impact of the loss of margin-injected regularization term. $\mathcal{R_\sigma}$ restricts MAGs to generate informative features that are more different with the original features (see Figure \ref{fig:augsimstats}(b)). In practice, Figure \ref{fig:augsimstats}(c) and (d) show that the features learned by our method (with margin-injected regularization) are more concentrated, e.g., the features of the same image are more similar and the gap between the features of the different images are enlarged, which proves informative augmented features can further lead the encoders to learn non-collapsed (scattered) features.

\begin{table*}[t]
	\tiny
	\renewcommand\arraystretch{1.1}
	\vskip -0.in
	\caption{Comparison of different methods on classification accuracy (top 1). We use \textit{conv} and \textit{fc} as backbones in the experiments. $^\ddagger$ denotes that the methods have reduced learnable parameters (See Appedix \ref{sec:networkarch}).}
	\vskip 0.in
	\label{tab:a}
	\setlength{\tabcolsep}{2.5pt}
	\begin{center}
		\begin{small}
			\begin{tabular}{l|cc||cc||cc||cc}
				\hline
				\multirow{2}*{Model} & \multicolumn{2}{c||}{Tiny ImageNet} & \multicolumn{2}{c||}{STL-10} & \multicolumn{2}{c||}{CIFAR10} & \multicolumn{2}{c}{CIFAR100} \\ 
				\cline{2-9}
				& conv & fc & conv & fc& conv & fc& conv & fc \\
				\hline
				\text{Fully supervised} & \multicolumn{2}{c||}{36.60} & \multicolumn{2}{c||}{68.70} & \multicolumn{2}{c||}{75.39} & \multicolumn{2}{c}{42.27} \\
				\hline
				\text{BiGAN} & 24.38 & 20.21 & 71.53 & 67.18 & 62.57 & 62.74 & 37.59 & 33.34 \\
				\text{NAT} & 13.70 & 11.62 & 64.32 & 61.43 & 56.19 & 51.29 & 29.18 & 24.57 \\
				\text{DIM} & 33.54 & 36.88 & 72.86 & 70.85 & 73.25 & 73.62 & 48.13 & 45.92 \\
				\text{SplitBrain$^{\ddagger}$} & 32.95 & 33.24 & 71.55 & 63.05 & 77.56 & 76.80 & 51.74 & 47.02 \\
				\text{SwAV} & 39.56 $\pm$ 0.2 & 38.87 $\pm$ 0.3 & 70.32 $\pm$ 0.4 & 71.40 $\pm$ 0.3 & 68.32 $\pm$ 0.2 & 65.20 $\pm$ 0.3 & 44.37 $\pm$ 0.3 & 40.85 $\pm$ 0.3 \\
				\text{SimCLR} & 36.24 $\pm$ 0.2 & 39.83 $\pm$ 0.1 & 75.57 $\pm$ 0.3 & 77.15 $\pm$ 0.3 & 80.58 $\pm$ 0.2 & 80.07 $\pm$ 0.2 & 50.03 $\pm$ 0.2 & 49.82 $\pm$ 0.3 \\
				\text{CMC$^{\ddagger}$} & 41.58 $\pm$ 0.1 & 40.11 $\pm$ 0.2 & 83.03 & 85.06 & 81.31 $\pm$ 0.2 & 83.28 $\pm$ 0.2 & 58.13 $\pm$ 0.2 & 56.72 $\pm$ 0.3 \\
				\text{MoCo} & 35.90 $\pm$ 0.2 & 41.37 $\pm$ 0.2 & 77.50 $\pm$ 0.2 & 79.73 $\pm$ 0.3 & 76.37 $\pm$ 0.3 & 79.30 $\pm$ 0.2 & 51.04 $\pm$ 0.2 & 52.31 $\pm$ 0.2 \\
				\text{BYOL} & 41.59 $\pm$ 0.2 & 41.90 $\pm$ 0.1 & 81.73 $\pm$ 0.3 & 81.57 $\pm$ 0.2 & 77.18 $\pm$ 0.2 & 80.01 $\pm$ 0.2 & 53.64 $\pm$ 0.2 & 53.78 $\pm$ 0.2 \\
				\text{Barlow Twins} & 39.81 $\pm$ 0.3 & 40.34 $\pm$ 0.2 & 80.97 $\pm$ 0.3 & 81.43 $\pm$ 0.3 & 76.63 $\pm$ 0.3 & 78.49 $\pm$ 0.2 & 52.80 $\pm$ 0.2 & 52.95 $\pm$ 0.2 \\
				\text{DACL} & 40.61 $\pm$ 0.2 & 41.26 $\pm$ 0.1 & 80.34 $\pm$ 0.2 & 80.01 $\pm$ 0.3 & 81.92 $\pm$ 0.2 & 80.87 $\pm$ 0.2 & 52.66 $\pm$ 0.2 & 52.08 $\pm$ 0.3 \\
				\text{LooC} & 42.04 $\pm$ 0.1 & 41.93 $\pm$ 0.2 & 81.92 $\pm$ 0.2 & 82.60 $\pm$ 0.2 & 83.79 $\pm$ 0.2 & 82.05 $\pm$ 0.2 & 54.25 $\pm$ 0.2 & 54.09 $\pm$ 0.2 \\
				\text{SimCLR + Debiased} & 38.79 $\pm$ 0.2 & 40.26 $\pm$ 0.2 & 77.09 $\pm$ 0.3 & 78.39 $\pm$ 0.2 & 80.89 $\pm$ 0.2 & 80.93 $\pm$ 0.2 & 51.38 $\pm$ 0.2 & 51.09 $\pm$ 0.2
				\\
				\text{SimCLR + Hard} & 40.05 $\pm$ 0.3 & 41.23 $\pm$ 0.2 & 79.86 $\pm$ 0.2 & 80.20 $\pm$ 0.2 & 82.13 $\pm$ 0.2 & 82.76 $\pm$ 0.1 & 52.69 $\pm$ 0.2 & 53.13 $\pm$ 0.2 \\
				\text{CMC$^{\ddagger}$ + Debiased} & 41.64 $\pm$ 0.2 & 41.36 $\pm$ 0.1 & 83.79 $\pm$ 0.3 & 84.20 $\pm$ 0.2 & 82.17 $\pm$ 0.2 & 83.72 $\pm$ 0.2 & 58.48 $\pm$ 0.2 & 57.16 $\pm$ 0.2 \\
				\text{CMC$^{\ddagger}$ + Hard} & 42.89 $\pm$ 0.2 & 42.01 $\pm$ 0.2 & 83.16 $\pm$ 0.3 & 85.15 $\pm$ 0.2 & 83.04 $\pm$ 0.2 & 86.22 $\pm$ 0.2 & 58.97 $\pm$ 0.3 & 59.13 $\pm$ 0.2 \\
				\hline
				\textbf{MetAug (only OUCL)$^{\ddagger}$} & 42.02 $\pm$ 0.1 & 42.14 $\pm$ 0.2 & 84.09 $\pm$ 0.2 & 84.72  $\pm$ 0.3 & 85.98 $\pm$ 0.2 & 87.13 $\pm$ 0.2 & 59.21 $\pm$ 0.2 & 58.73 $\pm$ 0.2 \\
				\textbf{MetAug$^{\ddagger}$} & \textbf{44.51 $\pm$ 0.2} & \textbf{45.36 $\pm$ 0.2} & \textbf{85.41 $\pm$ 0.3} & \textbf{85.62 $\pm$ 0.2} & \textbf{87.87 $\pm$ 0.2} & \textbf{88.12 $\pm$ 0.2} & \textbf{59.97 $\pm$ 0.3} & \textbf{61.06 $\pm$ 0.2} \\
				\hline
			\end{tabular}
		\end{small}
	\end{center}
	\vspace{-0.7cm}
\end{table*}

\begin{algorithm}[t]
	\vskip 0.in
	\begin{algorithmic}
		\STATE {\bfseries Input:} Multi-view dataset ${X}$ with $M$ views of each sample, minibatch size $n$, and hyperparameters $\alpha$, $\beta$, $\delta$.\\
		\STATE {\bf Initialize} The neural network parameters: ${\theta}$ and $\vartheta$ for view-specific encoders $f_{\theta}(\cdot)$ and projection heads $g_{\vartheta}(\cdot)$, $\omega$ for MAGs, i.e., $a_{\omega}(\cdot)$. The learning rates: $\ell$ and $\ell^\prime$.
		\REPEAT
		\FOR{$t$-th training iteration}
		\STATE Iteratively sample minibatch $\widetilde{X} = \left\{ {X_i} \right\}_{i = (t-1)n}^{tn}$.
		\STATE $\# \ regular \ contrastive \ training \ step$
		\STATE $\theta \leftarrow \theta - \ell\Delta_{\theta} \mathcal{L}_{MetAug} ( f_{\theta}, g_{\vartheta}, a_{\omega}, \widetilde{X} ) $ \\
		\STATE $\vartheta \leftarrow \vartheta - \ell\Delta_{\vartheta} \mathcal{L}_{MetAug} ( f_{\theta}, g_{\vartheta}, a_{\omega}, \widetilde{X} )$ \\
		\ENDFOR
		\FOR{$t$-th training iteration}
		\STATE Iteratively sample minibatch $\widetilde{X} = \left\{ {X_i} \right\}_{i = (t-1)n}^{tn}$.
		\STATE $\# \ compute \ fast \ weights$
		\STATE $\# \ retain \ computational \ graph$
		\STATE $\mathring{\theta} = \theta - \ell\Delta_{\theta} \mathcal{L}_{MetAug} ( f_{\theta}, g_{\vartheta}, a_{\omega}, \widetilde{X} ) $ \\
		\STATE $\mathring{\vartheta} = \vartheta - \ell\Delta_{\vartheta} \mathcal{L}_{MetAug} ( f_{\theta}, g_{\vartheta}, a_{\omega}, \widetilde{X} )$ \\
		\STATE $\# \ meta \ training \ step \ using \ second \ derivative$
		\STATE $\omega \leftarrow \omega - \ell^\prime\Delta_{\omega} \left( \mathcal{L}_{MetAug} ( f_{\mathring{\theta}}, g_{\mathring{\vartheta}}, a_{\omega}, \widetilde{X} ) + \alpha \cdot \mathcal{R_\sigma} \right)$ \\
		\ENDFOR
		\UNTIL $\theta$, $\vartheta$, and $\omega$ converge.
	\end{algorithmic}
	\vskip -0.in
	\caption{MetAug}
	\label{alg:metaug}
\end{algorithm}

\subsection{Optimization-driven unified contrast}
We propose to jointly contrast all features (including the original features and the meta augmented features) in one gradient back-propagation step. Motivated by \cite{FaceNet2}, we introduce the following optimization-driven unified loss function to replace the conventional contrastive loss as follows:
\begin{equation}
	\mathcal{L}_{OUCL} = \left[\sum_{k^- = 1}^{K^-} d(\{z^-\}_{k^-}) - \sum_{k^+ = 1}^{K^+} d(\{z^+\}_{k^+}) + \lambda \right]_{+}
	\label{eq:L}
\end{equation}%
where $[\cdot]_{+}$ ensures that $\mathcal{L} \geq 0$ is always held. Note that all original features and augmented features are involved. $\lambda$ is a margin between the summarized instance similarities to enhance the capability of the similarity separation. However, we find that the difference between $\sum_{k^- = 1}^{K^-} d(\{z^-\}_{k^-})$ and $\sum_{k^+ = 1}^{K^+} d(\{z^+\}_{k^+})$ is not the larger the better. Excessive increases of the difference may undermine the convergence in optimization. We thereby wish to adopt a margin $\lambda$ that leads to preferable convergence. We reform the loss in Equation \ref{eq:L} by adding a temperature coefficient $\beta$ as follows:
\begin{equation}
	\begin{aligned}
	\mathcal{L}_{OUCL} = &\frac{1}{\beta}log\Bigg\{1 + \sum_{k^- = 1}^{K^-} \sum_{k^+ = 1}^{K^+}\\ &exp\bigg[\beta\Big(d(\{z^-\}_{k^-}) - d(\{z^+\}_{k^+}) + \lambda\Big)\bigg]\Bigg\}
	\label{eq:LF}
	\end{aligned}
\end{equation}%
when $\beta \to +\infty$, Equation \ref{eq:LF} is Equation \ref{eq:L}. Inspired by \cite{Circle2}, we use weighting factors $\Gamma^-$ and $\Gamma^+$ to modulate the impacts of $d(\{z^-\}_{k^-})$ and $d(\{z^+\}_{k^+})$. Such approach aims to give greater weight to the similarity that deviates from the optimum and smaller weight to the similarity that has the close proximity with the optimum. $\Gamma^- = [d(\{z^-\}_{k^-})-O^-]_{+}$ and $\Gamma^+ = [O^+-d(\{z^+\}_{k^+})]_{+}$, where $O^-$ and $O^+$ represents the \textit{expected} optimums of $d(\{z^-\}_{k^-})$ and $d(\{z^+\}_{k^+})$. Note that, we further propose a \textit{variation} of $\Gamma$ including $\Gamma^-$ and $\Gamma^+$, and the comparisons of them are demonstrated in Section \ref{sec:gammavariant}. $\gamma^+$ and $\gamma^-$ is used to replace $\lambda$ and add $\Gamma^-$ and $\Gamma^+$ in Equation \ref{eq:LF}:
\begin{equation}
	\begin{aligned}
	&\mathcal{L}_{OUCL} = \frac{1}{\beta}log\Bigg\{1+\sum_{k^- = 1}^{K^-} \sum_{k^+ = 1}^{K^+}exp\bigg[\beta\Big(\Gamma^- \\& (d(\{z^-\}_{k^-})-\gamma^-)- \Gamma^+(d(\{z^+\}_{k^+})-\gamma^+)\Big)\bigg]\Bigg\}.
	\label{eq:LFa2}
	\end{aligned}
\end{equation}

We limit $d(\{z^-\}_{k^-})$ and $d(\{z^+\}_{k^+})$ in the range of $[0, 1]$ by normalizing the features in $\{z^-\}_{k^-}$ and $\{z^-\}_{k^-}$, such that \textit{theoretically}, the optimum of $d(\{z^-\}_{k^-})$ is $0$, the optimum of $d(\{z^+\}_{k^+})$ is $1$. The positive of $d(\{z^-\}_{k^-})-O^-$ and $O^+-d(\{z^+\}_{k^+})$ can easily be guaranteed. To cut the number of hyperparameters, we reform Equation \ref{eq:LFa2} into
\begin{equation}
	\begin{aligned}
	&\mathcal{L}_{OUCL} = \frac{1}{\beta}log\Bigg\{1+\sum_{k^- = 1}^{K^-} \sum_{k^+ = 1}^{K^+}exp\bigg[ \\& \beta\Big({(d(\{z^+\}_{k^+})-1)}^2+{(d(\{z^-\}_{k^-}))}^{2}-2\gamma^{2}\Big)\bigg]\Bigg\}
	\label{eq:Luni}
	\end{aligned}
\end{equation}
which is derived by setting $O^+=1+\gamma$, $O^-=-\gamma$, $\gamma^+=1-\gamma$, and $\gamma^-=\gamma$.

\subsection{Model objective}
Concretely, we adopt margin-injected meta feature augmentation in the contrastive learning paradigm to achieve desired discriminative multi-view representations, and the proposed $\mathcal{L}_{MetAug}$ is incorporated to replace the conventional contrastive loss $\mathcal{L}$. The final model objective is defined as:
\begin{equation}
	\mathcal{L}_{MetAug} = \mathcal{L}_{OUCL}^{ori} + \delta \cdot \mathcal{L}_{OUCL}^{aug}
	\label{equal16}
\end{equation}
where $\mathcal{L}_{OUCL}^{ori}$ represents the loss \textit{NOT} including the meta augmented features, $\mathcal{L}_{OUCL}^{aug}$ represents the loss including such features, and $\delta$ is a coefficient that controls the balance between them (we perform parameter comparisons in Appendix \ref{sec:paracomp}). It is worthy to note that the margin-injected regularization $\mathcal{R_\sigma}$ is only used in meta training the MAGs, i.e., $a_\omega(\cdot)$, while in regular training of encoders and projection heads, $\mathcal{R_\sigma}$ is discarded. $\mathcal{R_\sigma}$ restricts the augmented features to be informative so that such features can lead the encoder to efficiently and effectively learn discriminative representations. The training process is detailed by Algorithm \ref{alg:metaug}.

%% file: experiments.tex
\section{Experiments}\label{sec:experiments}
We benchmark our MetAug on five established datasets: Tiny ImageNet \cite{krizhevsky2009learning}, STL-10 \cite{coates2011analysis}, CIFAR10 \cite{krizhevsky2009learning}, CIFAR100 \cite{krizhevsky2009learning}, and ImageNet \cite{2009Feifei}. The compared benchmark methods include: BiGAN \cite{donahue2016adversarial}, NAT \cite{bojanowski2017unsupervised}, DIM \cite{hjelm2018learning}, SplitBrain \cite{Splitp2}, CPC \cite{OJ2019Data}, SwAV \cite{2020Mathilde}, SimCLR \cite{chen2020simple}, CMC \cite{Tian2019Contrastive}, MoCo \cite{2020Kaiming}, SimSiam \cite{2020ExploringChen}, InfoMin Aug. \cite{2020WhatTian}, BYOL \cite{2020Bootstrap}, Barlow Twins \cite{2021Barlow}, DACL \cite{2021Vikas} LooC \cite{2021Tete}, Debiased \cite{2020Debiased}, Hard \cite{2020Hard}, and NNCLR \cite{2021Dwibedi}.

\subsection{Efficiently performing MetAug} \label{sec:comparisondownstream}
\textbf{Implementations.} To efficiently perform CL within a restricted amount of the inputs in training, we uniformly set the batch size as 64 (see Appendix \ref{sec:batch} for the comparisons under different setting of batch size). For the experiments with \textit{conv} and \textit{fc} as the backbone networks, we adopt a network with the 5 convolutional layers in AlexNet \cite{2012Krizhevsky} as conv and a network with further 2 fully connected layers as fc. Inspired by the backbone splitting setting of SplitBrain \cite{Splitp2}, we evenly split the AlexNet into sub-networks across the channel dimension, and each sub-network is the view-specific encoder (see Appendix \ref{sec:implementation} for the detailed implementation). For the experiments with ResNet-50, we directly change the encoder network to ResNet-50. All backbone encoders are not pre-trained. MetAug (\textit{only OUCL}) is the ablation variant without margin-injected meta feature augmentation.

Given an RGB image, we convert it to the Lab image color space and split it into L and ab channels. During contrastive learning, RGB, L, and ab are used as three views of the image. Before feeding the views into our model, we simply adopt the same data augmentations in CMC \cite{Tian2019Contrastive}. Especially, the major contribution of DACL is the proposed data augmentation (i.e., mixup-noise) so that we particularly add mixup data augmentation for DACL. In training, a memory bank \cite{un2} is adopted to facilitate calculations. We retrieve 4096 past features from the memory bank to derive negatives. The learning rates and weight decay rates are uniform over comparisons.

\begin{table}[t]
	\small
	\renewcommand\arraystretch{1.1}
	\vskip -0.15in
	\caption{Performance (accuracy) on the CIFAR10 and STL-10 datasets with ResNet-50 \cite{2016Kaiming}.}
	\vskip -0.in
	\label{tab:b}
	\setlength{\tabcolsep}{5.5pt}
	\begin{center}
		\begin{tabular}{l|c|c|c}
			\hline
			\text{Model} & CIFAR10 & STL-10 & Average \\ 
			\hline
			\text{SwAV} & 83.15 & 82.93 & 83.04 \\
			\text{SimCLR} & 84.63 & 83.75 & 84.19 \\
			\text{CMC} & 86.10 & 86.83 & 86.47 \\
			\text{BYOL} & 87.14 & 87.56 & 87.35 \\
			\text{Barlow Twins} & 85.84 & 86.02 & 85.93 \\
			\text{DACL} & 86.93 & 88.11 & 87.52 \\
			\text{LooC} & 87.80 & 88.62 & 88.21 \\
			\text{SwAV + Hard} & 83.99 & 84.51 & 84.25
			\\
			\text{SimCLR + Hard} & 86.91 & 85.48 & 86.20 \\
			\text{CMC + Hard} & 88.25 & 87.79 & 88.02 \\
			\hline
			\textbf{MetAug (\textit{only OUCL})} & 88.79 & 88.31 & 88.55 \\
			\textbf{MetAug} & \textbf{91.09} & \textbf{90.26} & \textbf{90.68} \\
			\hline
		\end{tabular}
	\end{center}
	\vspace{-0.58cm}
\end{table}

%
%
\begin{table}[t]
	\vskip -0.1in
	\tiny
	\renewcommand\arraystretch{1.1}
	\caption{Linear evaluation results on ImageNet. We follow the setting of \cite{Tian2019Contrastive, 2021Dwibedi} to compare with other benchmark SSL methods with conv and ResNet-50. MetAug$^\ast$ denotes MetAug further leveraging the advances of the positive re-sampling technique \cite{2021Dwibedi}. Note that the batch size adopted by the compared methods are inconsistent in the comparisons using ResNet-50, i.e., CMC adopts 128 yet others adopt 4096. Therefore, we report the comparisons of MetAug using ResNet-50 with the batch size of 128 in Appendix \ref{sec:rescmc}.}
	\vskip -0.in
	\label{tab:c}
	\setlength{\tabcolsep}{8.5pt}
	\begin{center}
		\begin{small}
			\begin{tabular}{l|c|c|c}
				\hline
				\multicolumn{4}{c}{ImageNet} \\ 
				\cline{1-4} \multirow{2}*{Model} & conv & \multicolumn{2}{c}{ResNet-50} \\
				\cline{2-4}
				& \multicolumn{2}{c|}{top 1} & top 5 \\
				\hline
				\text{Fully supervised} & 50.5 & - & - \\
				\hline
				\text{SplitBrain} & 32.8 & - & - \\
				\text{CPC v2} & - & 63.8 & 85.3 \\
				\text{SwAV} & 38.0 $\pm$ 0.3 & 71.8 & - \\
				\text{SimCLR} & 37.7 $\pm$ 0.2 & 71.7 & - \\
				\text{CMC} & 42.6 & - & - \\
				\text{MoCo} & 39.4 $\pm$ 0.2 & 71.1 & - \\
				\text{SimSiam} & - & 71.3 & - \\
				\text{InfoMin Aug.} & - & 73.0 & 91.1 \\
				\text{BYOL} & 41.1 $\pm$ 0.2 & 74.3 & 91.6 \\
				\text{Barlow Twins} & 39.6 $\pm$ 0.2 & - & - \\
				\text{NNCLR} & - & 75.4 & 92.3 \\
				\text{DACL} & 41.8 $\pm$ 0.2 & - & - \\
				\text{LooC} & 43.2 $\pm$ 0.2 & - & - \\
				\text{SimCLR + Debiased} & 38.9 $\pm$ 0.3 & - & - \\
				\text{SimCLR + Hard} & 41.5 $\pm$ 0.2 & - & - \\
				\hline
				\textbf{MetAug} & \textbf{45.1 $\pm$ 0.2} & - & - \\
				\textbf{MetAug$^\ast$} & - & \textbf{76.0} & \textbf{93.2} \\
				\hline
			\end{tabular}
		\end{small}
	\end{center}
	\vskip -0.18in
\end{table}

\textbf{Comparison on downstream tasks.} We collect the results of 20 trials for comparisons. The average result of the last 20 epochs is used as the final result of each trial, and the 95\% confidence intervals are also reported, while the results without 95\% confidence intervals are quoted from the published papers. We compare MetAug against a fully-supervised method (similar to AlexNet \cite{2012Krizhevsky}) and the state-of-the-art unsupervised methods. Table \ref{tab:a} shows the comparisons on four benchmark datasets. The last two rows of tables represent the results of our methods. As demonstrated in tables, MetAug beats the best prior methods on all datasets. Even compared with the fully-supervised method trained end-to-end (without fine-tuning) for the architecture presented, the proposed method has a significant improvement on most downstream tasks, which demonstrates that MetAug can better model discriminative information when supervision is insufficient (e.g., the training data is limited). The ablation model (i.e., MetAug (\textit{only OUCL})) outperforms most unsupervised methods but falls short of the performance of MetAug. Thus, the ablation study proves the effectiveness of our proposed margin-injected meta feature augmentation and optimization-driven unified contrast.

\begin{table*}[t]
	\caption{Comparison of applying benchmark SSL methods with different data augmentations by using the fc backbone on CIFAR10.}
	\vskip -0.in
	\label{tab:d}
	\setlength{\tabcolsep}{9.58pt}
	\begin{center}
		\begin{tabular}{c|cccccc|ccc}
			\hline
			\multirow{3}*{ID}& \multicolumn{6}{c|}{Data augmentations} & \multicolumn{3}{c}{Methods} \\
			\cline{2-10}
			& horizontal & \multirow{2}*{rotate} & random & random & color & \multirow{2}*{mixup} & \multirow{2}*{DACL} & \multirow{2}*{LooC} & \multirow{2}*{MetAug} \\
			& flip & & crop & grey & jitter & & & & \\
			\hline
			1 & $\checkmark$ & $\checkmark$ & & & & & - & 80.73 & \textbf{87.05} \\
			2 & & & $\checkmark$ & & & & - & 81.16 & \textbf{87.53} \\
			3 & & & & $\checkmark$ & & & - & 80.70 & \textbf{86.81} \\
			4 & & & & & $\checkmark$ & & - & 81.64 & \textbf{87.79} \\
			5 & $\checkmark$ & & $\checkmark$ & & & & - & 82.05 & \textbf{88.12} \\
			6 & & $\checkmark$ & & & $\checkmark$ & & - & 82.16 & \textbf{88.01} \\
			7 & $\checkmark$ & & $\checkmark$ & & & $\checkmark$ & 80.87 & 82.21 & \textbf{88.22} \\
			8 & $\checkmark$ & $\checkmark$ & $\checkmark$ & $\checkmark$ & $\checkmark$ & $\checkmark$ & 82.09 & 83.17 & \textbf{88.65} \\
			\hline
		\end{tabular}
	\end{center}
	\vskip -0.15in
\end{table*}

DACL and LooC propose to enhance contrastive learning from the perspective of \textit{data} augmentation, while MetAug improves contrastive learning from the perspective of \textit{feature} augmentation. The idea behind our method is simple but effective, since contrastive learning works directly on features, and the augmented images need one step of encoding to become features. The experimental results support that MetAug achieves better performance on benchmarks.

\textbf{Performing MetAug on ResNet.} We perform classification comparisons on the CIFAR10 and STL-10 by using ResNet-50. Table \ref{tab:b} shows that MetAug and the ablation variant outperform the compared methods, which indicates that MetAug has strong adaptability to different encoders.

\subsection{Benchmarking MetAug on ImageNet}
\textbf{Implementation.} To comprehensively understand the performance of our proposed MetAug, we conduct comparisons on ImageNet and make fair comparisons with benchmark methods. The backbone encoder is conv or ResNet-50, and the results are demonstrated in Table \ref{tab:c}. MetAug is a decoupled approach so that we can introduce MetAug in the learning paradigm of state-of-the-art to improve the performance, e.g, for the experiments using conv or ResNet-50, we perform MetAug in CMC or NNCLR, respectively.

\textbf{Results.} As shown in Table \ref{tab:c}, we find that MetAug can effectively promote the performance of benchmark methods in the comparisons using both conv and ResNet-50. The results support that our proposed meta feature augmentation can enable different encoders to model discriminative information even in the large-scale dataset.


\subsection{Is MetAug robust for data augmentation?} \label{sec:robustda}
To illustrate the impacts of different data augmentations, we conducted multiple comparisons on CIFAR10 shown in Table \ref{tab:d}. Note that horizontal flip and rotate are similar, and we use them together in the 1-th comparison. In the 5-th comparison, we take the same data augmentations as the setting of comparisons in Section \ref{sec:comparisondownstream}. The data augmentations adopted in the 6-th comparison are as same as the setting of LooC \cite{2021Tete}. Additionally, \textit{mixup} is proposed by DACL \cite{2021Vikas}.

We observe from Table \ref{tab:d} that MetAug outperforms the compared methods in all comparisons. It is worth noting that even using weak data augmentation degenerates the performance of our method as well as benchmark methods, but the performance degeneration of our method is minimal compared to others, e.g., from 8-th and 1-th comparison, we find that the gap of MetAug is 1.60\%, while that of LooC is 2.44\%. The results support that MetAug is robust for various data augmentations.

\begin{figure}
	\vskip 0.in
	\centering
	\includegraphics[width=0.48\textwidth]{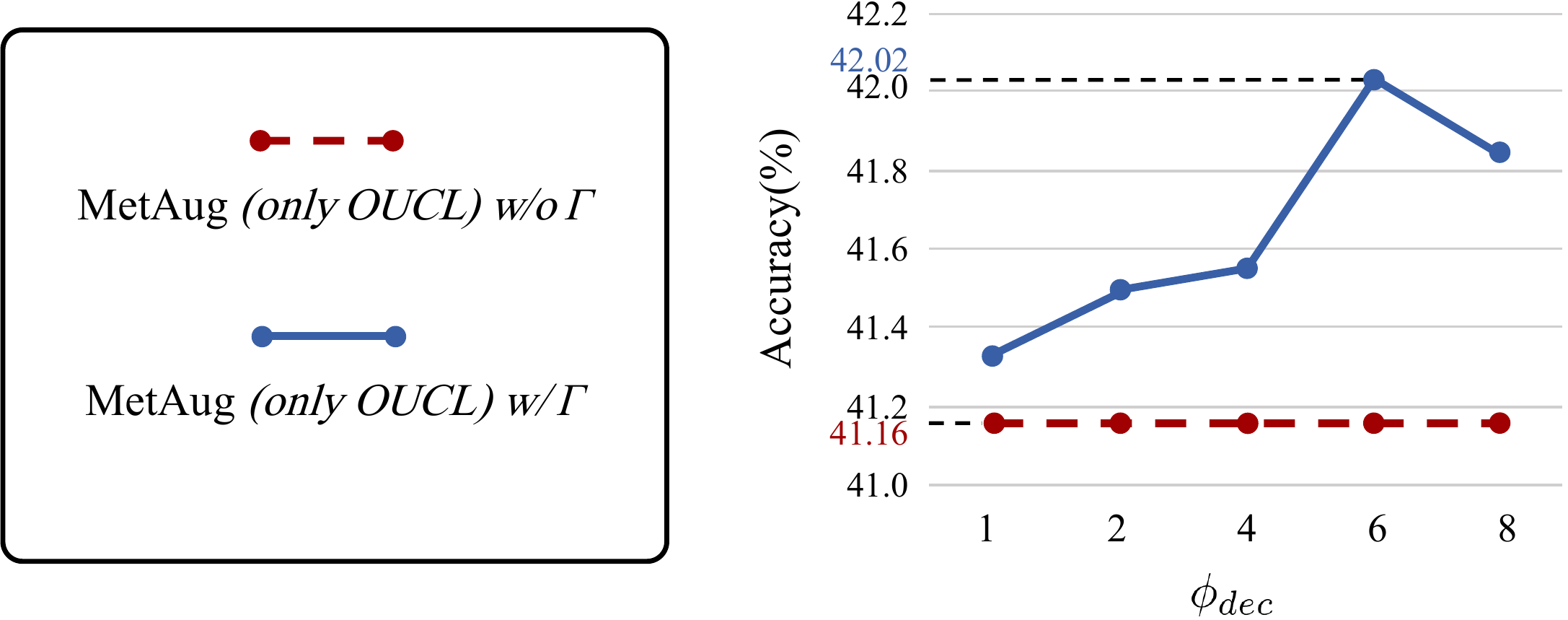}
	\vskip -0.15in
	\caption{The impact of different $\phi_{dec}$ on the performance of our method using $\bar{\Gamma}$. Comparisons are conducted on Tiny ImageNet with conv as the encoders.}
	\label{fig:leveragingfactor}
	\vspace{-0.6cm}
\end{figure}

\subsection{Do the variant of $\Gamma$ promote MetAug?} \label{sec:gammavariant}
In practice, we find that the introduction of the weighting factor $\Gamma$ cannot directly improve our proposed method. Our conjuncture lies in that $\Gamma$ may cause the loss to converge excessively fast, which leaves the network parameters at a local minimum. Therefore, we propose a variant to replace $\Gamma$ in Equation \ref{eq:LFa2}, i.e., $\bar{\Gamma} = \dfrac{\Gamma}{\phi_{dec}}$ where $\phi_{dec}$ is a linear attenuation coefficient to linearly attenuate the impact of $\Gamma$ so that the difference between the current value and the optimum becomes smaller.

We use MetAug (\textit{only OUCL}) to demonstrate the effectiveness of the proposed variant. The results are shown in Figure \ref{fig:leveragingfactor}. We observe that the performance of our method get peak value when $\phi_{dec}$ is 6, which manifests that introducing a certain linear attenuation to $\Gamma$ can promote MetAug.

%% file: conclusion.tex
\section{Conclusion} \label{sec:conclusion}
We conclude that exploring informative features is the key to contrastive learning. Different from the conventional contrastive methods that collect enough informative features to learn a good representation by enlarging the batch or memory bank, we motivate MetAug to learn a discriminative representation from a restricted amount of images. Our method proposes margin-injected meta feature augmentation to straightforwardly augment features to be informative and avoid learning degenerate features. To efficiently make use of all available features, MetAug further proposes optimization-driven unified contrast. Experimental evaluations demonstrate that MetAug achieves the state-of-the-art.

\section*{Acknowledgements}
The authors would like to thank the anonymous reviewers for their valuable comments. This work is supported in part by National Natural Science Foundation of China No. 61976206 and No. 61832017, Key Special Project for Introduced Talents Team of Southern Marine Science and Engineering Guangdong Laboratory (Guangzhou) No. GML2019ZD0603, National Key Research and Development Program of China No. 2019YFB1405100, Beijing Outstanding Young Scientist Program NO. BJJWZYJH012019100020098, Beijing Academy of Artificial Intelligence (BAAI), the Fundamental Research Funds for the Central Universities, the Research Funds of Renmin University of China 21XNLG05, and Public Computing Cloud, Renmin University of China. This work is also supported in part by Intelligent Social Governance Platform, Major Innovation \& Planning Interdisciplinary Platform for the “Double-First Class” Initiative, Renmin University of China, and Public Policy and Decision-making Research Lab of Renmin University of China.


%% file: appendix.tex
\clearpage
\appendix
\section{Appendix - Extended comparisons}

\begin{figure}
	\vskip 0.in
	\centering
	\includegraphics[width=0.45\textwidth]{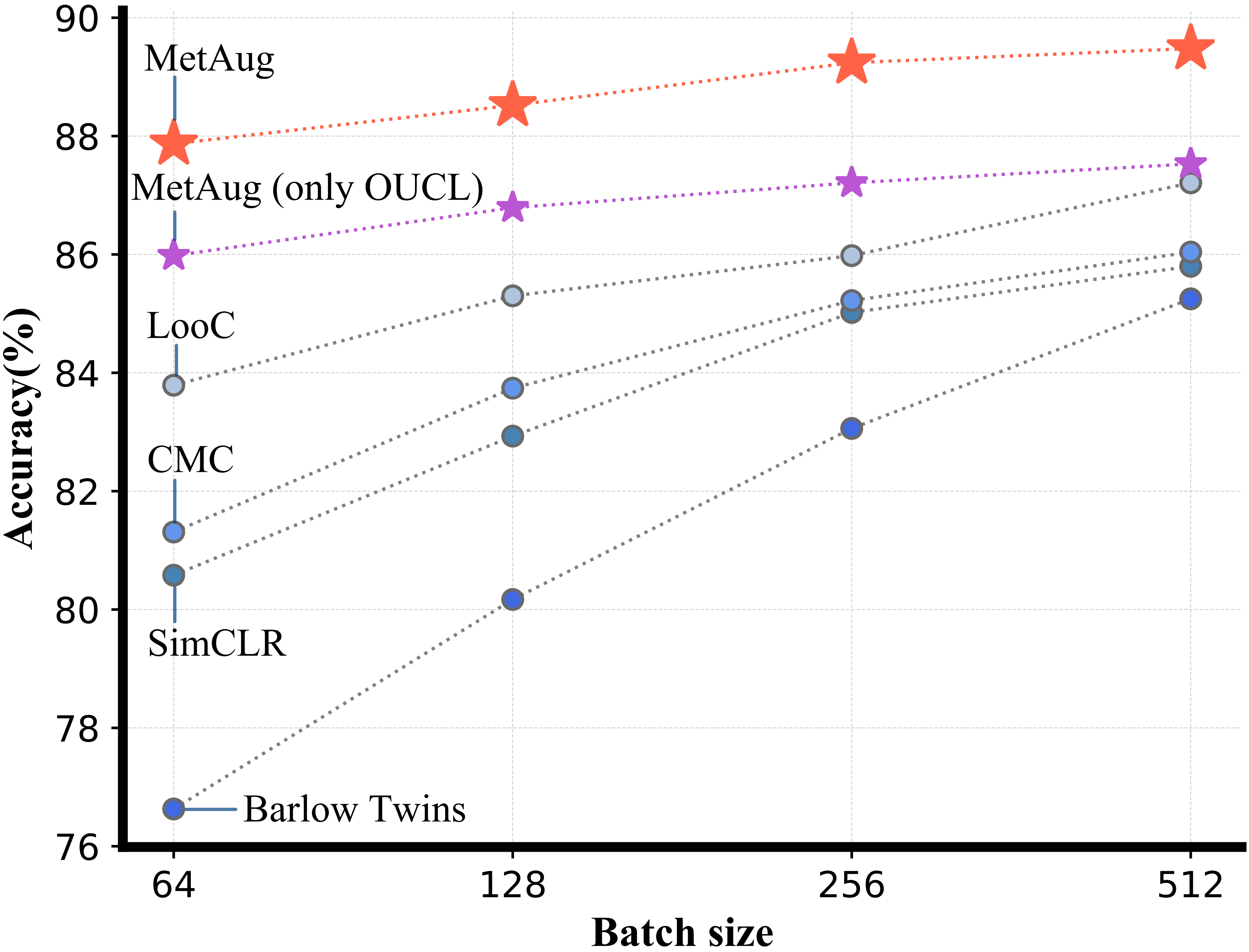}
	\vskip -0.1in
	\caption{Comparison of different methods on classification accuracy (top1) under various settings of batch size. We conducted experiments on CIFAR10 with conv encoder.}
	\label{fig:batchaccstats}
	\vspace{-0.3cm}
\end{figure}

In this section, we provide several experimental analyses about the advantages of our proposed method. The experiments to find appropriate hyperparameters are conducted as well, and in detail, we conduct comparisons of using different hyperparameters on the validation set of corresponding benchmark datasets.

\subsection{Can MetAug perform consistently under different settings of batch size?} \label{sec:batch}
As the results shown in Table \ref{tab:a}, \ref{tab:b}, and \ref{tab:c}, we observe that MetAug achieves our expectation that learning anti-collapse and discriminative representations from a restricted amount of images in a training step (i.e., the batch size is limited). However, we conduct further experiments to explore whether MetAug has consistent performance under settings of larger batch sizes.

From Figure \ref{fig:batchaccstats}, we observe that with the increase of batch size, each compared method achieves better performance on the downstream task. We conjecture that with the enlarging of batch size, the number of available features in a training step is increased, so that models may explore more informative features to promote the performance of contrastive learning. Yet, comparing our method with the benchmark methods, we find that the gap between the performance of MetAug (\textit{only OUCL}) and the compared methods becomes smaller. We extend the mentioned conjecture: as more informative features can be explored by all methods in a training step, OUCL’s advantage becomes less significant. OUCL aims to include all available features to efficiently train the model and avoid the optimization fall into a local optimum, and the increase of batch size, which means sufficient self-supervision, can naturally promote the efficiency of optimization and avoid the fall into a local optimum. Yet the advance of OUCL is always maintained, which is supported by the comparison. Only LooC's performance can gradually catch up with the performance of MetAug (\textit{only OUCL}). We research the setting of LooC, and find that LooC leverages more than one (e.g., three) contrastive loss in a training step, which allows LooC to train the model multiple times. We observe that, even in the large batch size, MetAug can still improve the state-of-the-art methods by a significant margin.

Concretely, MetAug maintains its superiority over compared method under different settings of batch size.

\begin{figure}
	\vskip 0.in
	\centering
	\includegraphics[width=0.4\textwidth]{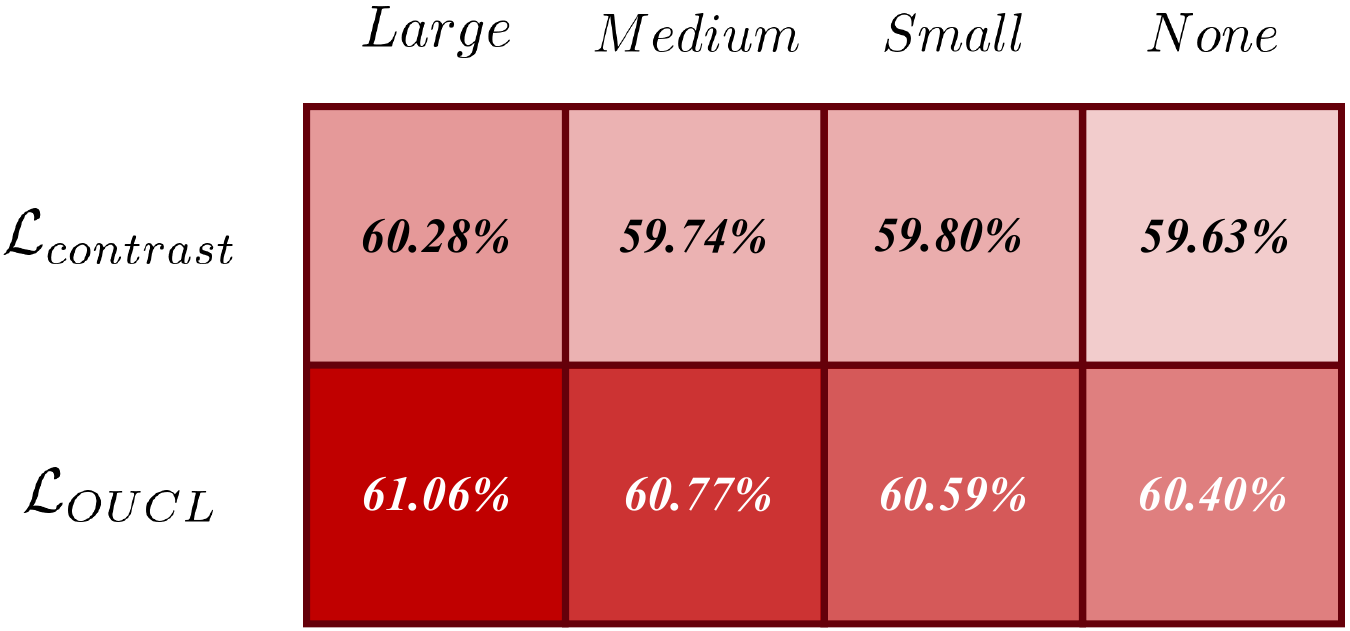}
	\vskip -0.1in
	\caption{Heatmap of injected margin variant comparisons.}
	\label{fig:marginvariants}
	\vspace{-0.6cm}
\end{figure}

\begin{table}[t]
	\small
	\renewcommand\arraystretch{1.1}
	\vskip -0.in
	\caption{Performance (accuracy) of MetAug with or without the augmented features on CIFAR10 with conv encoder.}
	\vskip -0.in
	\label{tab:e}
	\setlength{\tabcolsep}{4.pt}
	\begin{center}
		\begin{tabular}{l|c|c|c}
			\hline
			\multirow{2}*\text{Model} & \multirow{2}*{$\delta$} & w/ augmented & w/o augmented \\ 
			& & features & features \\
			\hline
			\text{SimCLR} & - & \multicolumn{2}{c}{80.58} \\
			\text{DACL} & - & \multicolumn{2}{c}{81.92} \\
			\text{LooC} & - & \multicolumn{2}{c}{83.79} \\
			\text{CMC + Hard} & - & \multicolumn{2}{c}{83.04} \\
			\hline
			\multirow{8}*{\textbf{MetAug}} & $10^{-1}$ & \textbf{85.85} & 85.48 \\
			& $10^{-2}$ & 85.91 & \textbf{85.99} \\
			& $10^{-3}$ & 86.57 & \textbf{86.65} \\
			& $10^{-4}$ & \textbf{87.42} & 87.41 \\
			& $10^{-5}$ & 87.72 & \textbf{87.87} \\
			& $10^{-6}$ & 87.26 & \textbf{87.47} \\
			& $10^{-7}$ & 86.90 & \textbf{87.19} \\
			& $10^{-8}$ & 86.12 & \textbf{86.35} \\
			\hline
		\end{tabular}
	\end{center}
	\vspace{-0.88cm}
\end{table}

\begin{figure*}
	\vskip 0.in
	\centering
	\includegraphics[width=0.95\textwidth]{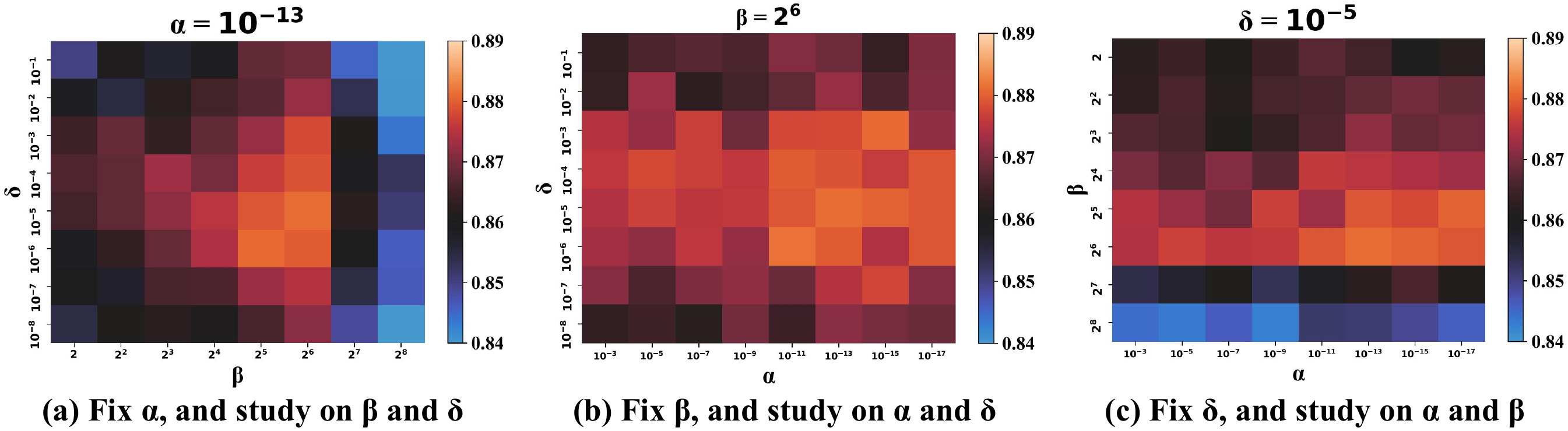}
	\vskip -0.1in
	\caption{Impacts of the hyperparameters $\alpha$, $\beta$, and $\delta$ of our proposed method. We conducted comparisons based on MetAug on CIFAR10 with fc encoder. To measure the influences, we iteratively fixed one parameter and then study on the others by selecting them in the ranges, respectively.}
	\label{fig:heatmap}
	\vspace{-0.3cm}
\end{figure*}

\begin{figure}
	\vskip 0.in
	\centering
	\includegraphics[width=0.45\textwidth]{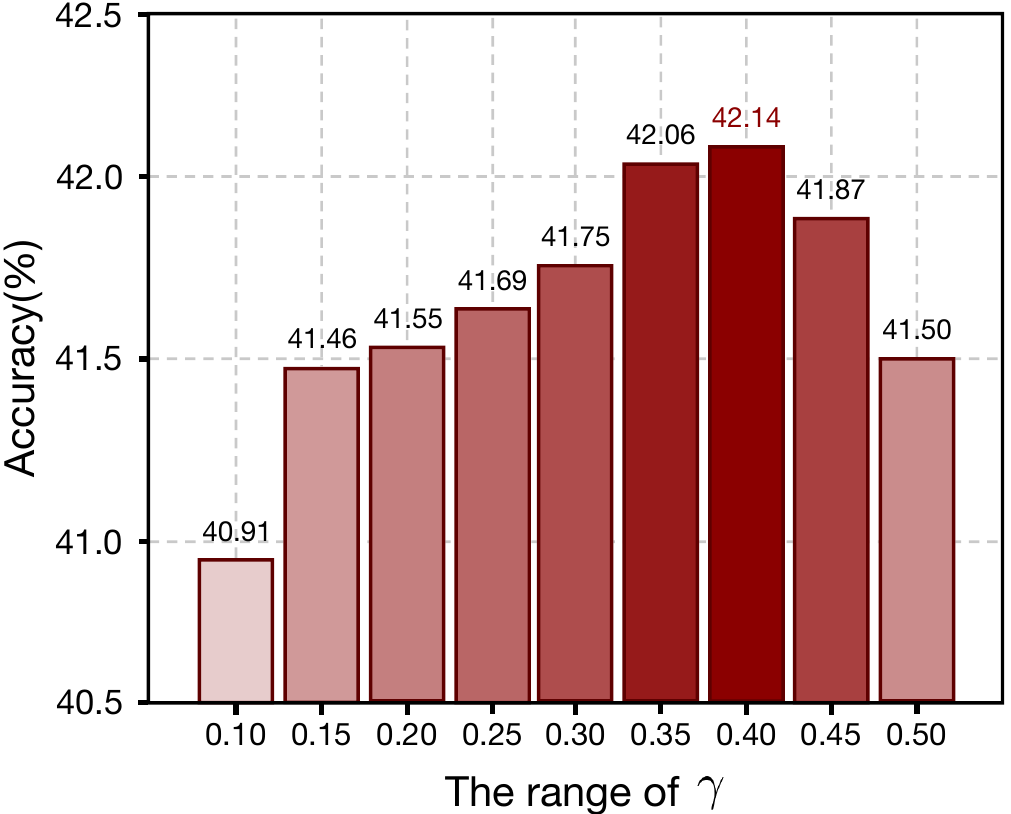}
	\vskip -0.1in
	\caption{Impacts of the hyperparameter $\gamma$ of our proposed method. We conducted comparisons based on MetAug (\textit{onlu OUCL}) on Tiny ImageNet with fc encoder. To measure the impact of $\gamma$, we iteratively select $\gamma$ and observe the accuracy of the method.}
	\label{fig:gammahpstats}
	\vspace{-0.5cm}
\end{figure}

\subsection{Variants of the injected margin} \label{sec:marginvariant}
We denote $\min^+$ as $\min(\{d(\{z^+\}_{k^+})\})$ and $\max^-$ as $\max(\{d(\{z^-\}_{k^-})\})$. For \textit{"Medium"}, both $\sigma^+$ and $\sigma^-$ equal to $mean[\min^+, \max^-]$. $\sigma^+ = \max[\min^+, \max^-]$ and $\sigma^- = \min[\min^+, \max^-]$ in \textit{"Small"}.

In Figure \ref{fig:marginvariants}, we conduct comparisons on CIFAR100 with fc. We observe that, whether our method uses $\mathcal{L}_{contrast}$ or the proposed $\mathcal{L}_{OUCL}$, all three variants can improve MetAug, and our method with \textit{"Large"} achieve the best performance. The experiments further prove the effectiveness of the two key ingredients of MetAug.

\subsection{Understanding of the augmented features} \label{sec:rethinkaug}
To understand the augmented features, we conduct a comparison of MetAug by adopting the augmented features in the test or not. As shown in Table \ref{tab:e}, the results of MetAug using the augmented features in the test are listed in the \textit{w/ augmented features} column, and the results of MetAug NOT using the augmented features in the test are listed in the \textit{w/o augmented features} column (which is the regular approach in the test). We select $\delta$ from the range of \{$10^{-1}, 10^{-2}, 10^{-3}, 10^{-4}, 10^{-5}, 10^{-6}, 10^{-7}, 10^{-8}$\} to generate different augmented features. Specifically, the approach of adopting the augmented features in the test is that we use MAGs to generate augmented features and such features are treated as the same as the original features, i.e., augmented features are regarded as additions of original features. Note that, in regular test (i.e., w/ augmented features), we use the representation $h_i^j$ and discard the projection head $g_{\vartheta_j}(\cdot)$ to feed into the classifier, while, in the test of using augmented features, we have to use the feature $z_i^j$ generated by the projection head $g_{\vartheta_j}(\cdot)$ to feed into the classifier, because MAGs work on the feature $z_i^j$.

From Table \ref{tab:e}, we observe that, generally, MetAug \textit{w/o augmented features} beats MetAug \textit{w/ augmented features}. The reasons behind such phenomenon are: 1) the augmented features are generated to lead the encoders to learn discriminative representations (e.g., $h_i^j$), which indicates that the augmented features contribute to the improvement of the encoders, but this does not mean that the augmented features are discriminative for downstream tasks; 2) in the test of using augmented features, we do not discard the projection head $g_{\vartheta_j}(\cdot)$, and recent works prove that the approach of using a projection head in training and discarding such head in the test can significantly improve the performance of the model on downstream tasks \cite{chen2020simple, 2020Kaiming}.

Out of the understanding of the experimental results, we think that the augmented features contain useful information that can improve the encoder, but such information may not be discriminative to downstream tasks.

\subsection{Synthetic comparison of hyperparameters} \label{sec:paracomp}
To intuitively understand the impacts of hyperparameters, we conduct comparisons by using various combinations of them for the proposed MetAug. Specifically, $\alpha$ controls the impact of the proposed margin-injected regularization term. The hyperparameter $\beta$ is proposed as a temperature coefficient in OUCL. $\gamma$ is a specific parameter to replace the hyperparameters in OUCL such that the number of hyperparameters can be reduced. $\delta$ balances the impact of OUCL that uses augmented features and OUCL that does not use these features.

As demonstrated in Figure \ref{fig:gammahpstats}, we first solely study $\gamma$'s impact on MetAug, because $\gamma$ is only used in OUCL function, and in practice, we find that, compared with other hyperparameters, $\gamma$ has less impact on our method. We conduct experiments on Tiny ImageNet with fc encoder and select $\gamma$ from the corresponding range for MetAug (\textit{only OUCL}) to clarify its impact, and the results indicate that appropriate selected $\gamma$ can indeed promote the performance of our method, but the differences between the impacts of different $\gamma$ are limited.

Then, we fix $\gamma = 0.40$ and study on the impacts of other hyperparameters. As the results are shown in Figure \ref{fig:heatmap}, the plots further elaborate our parameter studies' results with MetAug on the CIFAR10 benchmark dataset with fc encoder. To explore the influence of $\beta$ and $\delta$, we first fixed $\alpha=10^{-13}$, and then we selected $\beta$ from the range of \{$2, 2^{2}, 2^{3}, 2^{4}, 2^{5}, 2^{6}, 2^{7}, 2^{8}$\} and $\delta$ from the range of \{$10^{-1}, 10^{-2}, 10^{-3}, 10^{-4}, 10^{-5}, 10^{-6}, 10^{-7}, 10^{-8}$\}. Following the same experimental principle as above, we selected $\alpha$ from the range of \{$10^{-3}, 10^{-5}, 10^{-7}, 10^{-9}, 10^{-11}, 10^{-13}, 10^{-15}, 10^{-17}$\}. See Figure \ref{fig:heatmap}(a), (b), and (c) for the details of the comparisons. In general, good classification performance highly depends on the $\beta$ and $\delta$ terms. Also, $\alpha$ is an intensely necessary supplement for adapting the interval between similarities of augmented features and original features, which avoids to learn degenerate representations. We also find that the potential to improve the learned representations grows with the adjustment of term $\beta$, e.g., the initial loss becomes relatively large.

\begin{table}[t]
	\vskip -0.1in
	\tiny
	\renewcommand\arraystretch{1.}
	\caption{Comparisons on ImageNet with different view settings using ResNet-50 as the encoder. $\dagger$ denotes our reimplementations. RandAugment is proposed by \cite{DBLP:journals/corr/abs-1909-13719}.}
	\vskip -0.in
	\label{tab:imrescmc}
	\setlength{\tabcolsep}{4.pt}
	\begin{center}
		\begin{small}
			\begin{tabular}{c|l|c|c}
			    \hline
				\text{View} & \multirow{2}*{\text{Method}} & \multicolumn{2}{c}{ResNet-50} \\
				\cline{3-4}
				\text{setting} & & top 1 & top 5 \\
				\hline
				\hline
				\multirow{5}*{\rotatebox{90}{\textit{L, ab}}} & \text{CMC} & \textbf{64.0} & \textbf{85.5} \\
				& \text{CMC}$^\dagger$ & 63.3 & 84.8 \\
				\cline{2-4}
				& \textbf{MetAug (only OUCL)} & 63.9 & 85.7 \\
				& \textbf{MetAug (only MAG)} & 64.4 & 86.0 \\
				& \textbf{MetAug} & \boxed{\textbf{65.1}} & \boxed{\textbf{86.2}} \\
				\hline
				\hline
				\multirow{8}*{\rotatebox{90}{\textit{Y, Db, Dr}}} & \text{CMC} & 64.8 & 86.1 \\
				& \text{CMC}$^\dagger$ & 64.5 & 86.2 \\
				& \text{CMC + RandAugment} & \textbf{66.2} & \textbf{87.0} \\
				& \text{CMC + RandAugment}$^\dagger$ & 65.7 & 86.8 \\
				\cline{2-4}
				& \textbf{MetAug (only OUCL)} & 65.0 & 86.3 \\
				& \textbf{MetAug (only MAG)} & 65.9 & 86.8 \\
				& \textbf{MetAug} & 66.4 & 87.1 \\
				& \textbf{MetAug + RandAugment} & \boxed{\textbf{66.7}} & \boxed{\textbf{87.5}} \\
				\hline
			\end{tabular}
		\end{small}
	\end{center}
	\vskip -0.37in
\end{table}

\subsection{Discussion of the comparisons using ResNet-50 on ImageNet} \label{sec:rescmc}
In Table \ref{tab:c}, we do not report the experimental results of CMC using ResNet-50, because the batch sizes adopted by them are inconsistent, i.e., CMC adopts 128 yet other methods adopt 4096. Therefore, the performance of CMC is not competitive compared to other compared methods, including MetAug$^\ast$. Our proposed MetAug can be treated as a \textit{feature augmentation approach} so that MetAug can be embedded into each self-supervised learning architecture. The original MetAug is implemented based on CMC, so for a fair comparison, MetAug adopts the same batch size as CMC, while MetAug$^\ast$ is implemented based on NNCLR \cite{2021Dwibedi} with the batch size of 4096. Therefore, MetAug$^ast$ apparently outperforms MetAug, so we only report the best variant of our method, i.e., MetAug$^ast$. Whereas, we have comprehensively evaluated whether our method can really improve the performance of CMC on ImageNet, which is shown in Table \ref{tab:imrescmc}. MetAug can be treated as a component to improve baseline methods, which is implemented based on CMC$^\dagger$. We observe that MetAug beats CMC and even achieves better performance than CMC + RandAugment, which proves that our method can not only improve CMC but also beat RandAugment. We further employ RandAugment in our method. The results show that such a variant achieves the best performance, which is consistent with \ref{sec:robustda}, i.e., stronger data augmentation improves the performance of our method.

\section{Appendix - Implementation} \label{sec:implementation}
In this paper, we introduce a novel self-supervised representation learning approach, i.e., \textit{\textbf{Met}a Feature \textbf{Aug}mentation} (MetAug), of which Figure \ref{fig:algoframe} depicts the overview framework. The following subsections provide the design details of MetAug.

\subsection{Network architecture} \label{sec:networkarch}
In the experiments, neural network classification methods (i.e., conv and fc) are adopted as the backbone networks, and the classifiers (i.e., the linear networks) on the representations extracted from the encoders are performed on downstream classification tasks.

According to the principle of building the encoders, the AlexNet is split across the channel dimension with the idea that split-AlexNet can also perform well in learning representations between views, which only has the halved learnable parameters \cite{Splitp2}. We build the AlexNet with 5 convolutional layers, 2 linear layers, and a fully connected layer followed by a l2 normalization function. Then the split-AlexNets (i.e., the sub-networks) are regarded as the encoders. In experiments, we use conv and fc, which use the corresponding layers of AlexNet. Note that we split AlexNet across channels for RGB, L, and ab views. in the test, we concatenate representations layer-wise from the encoders into one to achieve the final representations of the inputs.

We develop the classifier by leveraging a linear network followed by a softmax output function. Following the proposed experimental setting of the previous literature \cite{2018RepresentationOord, hjelm2018learning, 2019Arora, Tian2019Contrastive}, we evaluate the quality of the learned representations by freezing the weights of backbone encoders and training the linear classifier in the test.

\subsection{Algorithm description} \label{sec:algocode}
MetAug is an end-to-end representation learning method: we iteratively train the encoders and MAGs by back-propagating $\mathcal{L}_{MetAug}$, and the training process is based on Adam gradient optimization.

The proposed MetAug is a generalized approach, which can be used for various downstream tasks, e.g., classification, clustering, regression, etc. We can straightforwardly train the encoders, pretrained by MetAug, on downstream tasks. The detailed implementation is available at \url{https://github.com/jiangmengli/metaug}.

%% file: main.bbl
\begin{thebibliography}{54}
\providecommand{\natexlab}[1]{#1}
\providecommand{\url}[1]{\texttt{#1}}
\expandafter\ifx\csname urlstyle\endcsname\relax
  \providecommand{\doi}[1]{doi: #1}\else
  \providecommand{\doi}{doi: \begingroup \urlstyle{rm}\Url}\fi

\bibitem[a et~al.(2002)a, Bengio, Bengio, Cloutier, and Gecsei]{Succ2002On}
a, S., Bengio, S., Bengio, Y., Cloutier, J., and Gecsei, J.
\newblock On the optimization of a synaptic learning rule.
\newblock 2002.

\bibitem[Arora et~al.(2019)Arora, Khandeparkar, Khodak, Plevrakis, and
  Saunshi]{2019Arora}
Arora, S., Khandeparkar, H., Khodak, M., Plevrakis, O., and Saunshi, N.
\newblock A theoretical analysis of contrastive unsupervised representation
  learning.
\newblock 2019.

\bibitem[Bachman et~al.(2019)Bachman, Hjelm, and Buchwalter]{2019Philip}
Bachman, P., Hjelm, R.~D., and Buchwalter, W.
\newblock Learning representations by maximizing mutual information across
  views.
\newblock In \emph{NeurIPS 2019}, 2019.

\bibitem[Bengio et~al.(2002)Bengio, Bengio, and Cloutier]{2002Bengio}
Bengio, Y., Bengio, S., and Cloutier, J.
\newblock Learning a synaptic learning rule.
\newblock In \emph{Ijcnn-91-seattle International Joint Conference on Neural
  Networks}, 2002.

\bibitem[Bojanowski \& Joulin(2017)Bojanowski and
  Joulin]{bojanowski2017unsupervised}
Bojanowski, P. and Joulin, A.
\newblock Unsupervised learning by predicting noise.
\newblock \emph{arXiv preprint arXiv:1704.05310}, 2017.

\bibitem[Caron et~al.(2020)Caron, Misra, Mairal, Goyal, Bojanowski, and
  Joulin]{2020Mathilde}
Caron, M., Misra, I., Mairal, J., Goyal, P., Bojanowski, P., and Joulin, A.
\newblock Unsupervised learning of visual features by contrasting cluster
  assignments.
\newblock 2020.

\bibitem[Chen et~al.(2020)Chen, Kornblith, Norouzi, and Hinton]{chen2020simple}
Chen, T., Kornblith, S., Norouzi, M., and Hinton, G.
\newblock A simple framework for contrastive learning of visual
  representations.
\newblock \emph{arXiv preprint arXiv:2002.05709}, 2020.

\bibitem[Chen \& He(2020)Chen and He]{2020ExploringChen}
Chen, X. and He, K.
\newblock Exploring simple siamese representation learning.
\newblock 2020.

\bibitem[Chen et~al.(2016)Chen, Hoffman, Colmenarejo, Denil, Lillicrap,
  Botvinick, and Freitas]{2016LearningChen}
Chen, Y., Hoffman, M.~W., Colmenarejo, S.~G., Denil, M., Lillicrap, T.~P.,
  Botvinick, M., and Freitas, N.~D.
\newblock Learning to learn without gradient descent by gradient descent.
\newblock 2016.

\bibitem[Chopra et~al.(2005)Chopra, HAdSell, and Lecun]{2005Chopra}
Chopra, S., HAdSell, R., and Lecun, Y.
\newblock Learning a similarity metric discriminatively, with application to
  face verification.
\newblock In \emph{2005 IEEE Computer Society Conference on Computer Vision and
  Pattern Recognition (CVPR'05)}, 2005.

\bibitem[Chuang et~al.(2020)Chuang, Robinson, Lin, Torralba, and
  Jegelka]{2020Debiased}
Chuang, C.~Y., Robinson, J., Lin, Y.~C., Torralba, A., and Jegelka, S.
\newblock Debiased contrastive learning.
\newblock 2020.

\bibitem[Coates et~al.(2011)Coates, Ng, and Lee]{coates2011analysis}
Coates, A., Ng, A., and Lee, H.
\newblock An analysis of single-layer networks in unsupervised feature
  learning.
\newblock In \emph{Proceedings of the fourteenth international conference on
  artificial intelligence and statistics}, 2011.

\bibitem[Cubuk et~al.(2019)Cubuk, Zoph, Shlens, and
  Le]{DBLP:journals/corr/abs-1909-13719}
Cubuk, E.~D., Zoph, B., Shlens, J., and Le, Q.~V.
\newblock Randaugment: Practical data augmentation with no separate search.
\newblock \emph{CoRR}, abs/1909.13719, 2019.
\newblock URL \url{http://arxiv.org/abs/1909.13719}.

\bibitem[Donahue et~al.(2016)Donahue, Kr{\"a}henb{\"u}hl, and
  Darrell]{donahue2016adversarial}
Donahue, J., Kr{\"a}henb{\"u}hl, P., and Darrell, T.
\newblock Adversarial feature learning.
\newblock \emph{arXiv preprint arXiv:1605.09782}, 2016.

\bibitem[du~Plessis et~al.(2014)du~Plessis, Niu, and Sugiyama]{2014Marthinus}
du~Plessis, M.~C., Niu, G., and Sugiyama, M.
\newblock Analysis of learning from positive and unlabeled data.
\newblock In \emph{Advances in Neural Information Processing Systems 27: Annual
  Conference on Neural Information Processing Systems 2014, December 8-13 2014,
  Montreal, Quebec, Canada}, 2014.

\bibitem[Dwibedi et~al.(2021)Dwibedi, Aytar, Tompson, Sermanet, and
  Zisserman]{2021Dwibedi}
Dwibedi, D., Aytar, Y., Tompson, J., Sermanet, P., and Zisserman, A.
\newblock With a little help from my friends: Nearest-neighbor contrastive
  learning of visual representations.
\newblock 2021.

\bibitem[Elkan \& Noto(2008)Elkan and Noto]{2008Charles}
Elkan, C. and Noto, K.
\newblock Learning classifiers from only positive and unlabeled data.
\newblock In \emph{Proceedings of the 14th {ACM} {SIGKDD} International
  Conference on Knowledge Discovery and Data Mining, Las Vegas, Nevada, USA,
  August 24-27, 2008}. {ACM}, 2008.

\bibitem[Ermolov et~al.(2020)Ermolov, Siarohin, Sangineto, and
  Sebe]{2020WhiteningErmolov}
Ermolov, A., Siarohin, A., Sangineto, E., and Sebe, N.
\newblock Whitening for self-supervised representation learning.
\newblock 2020.

\bibitem[Finn et~al.(2017)Finn, Abbeel, and Levine]{2017ModelFinn}
Finn, C., Abbeel, P., and Levine, S.
\newblock Model-agnostic meta-learning for fast adaptation of deep networks.
\newblock In \emph{Proceedings of the 34th International Conference on Machine
  Learning, {ICML} 2017, Sydney, NSW, Australia, 6-11 August 2017}, Proceedings
  of Machine Learning Research, pp.\  1126--1135. {PMLR}, 2017.

\bibitem[Goldberger et~al.(2003)Goldberger, Gordon, and
  Greenspan]{2003Goldbberger}
Goldberger, J., Gordon, S., and Greenspan, H.
\newblock An efficient image similarity measure based on approximations of
  kl-divergence between two gaussian mixtures.
\newblock In \emph{IEEE International Conference on Computer Vision}, 2003.

\bibitem[Grill et~al.(2020)Grill, Strub, Altché, Tallec, Richemond,
  Buchatskaya, Doersch, Pires, Guo, and Azar]{2020Bootstrap}
Grill, J.~B., Strub, F., Altché, F., Tallec, C., Richemond, P.~H.,
  Buchatskaya, E., Doersch, C., Pires, B.~A., Guo, Z.~D., and Azar, M.~G.
\newblock Bootstrap your own latent: A new approach to self-supervised
  learning.
\newblock 2020.

\bibitem[Gutmann \& Hyv{\"{a}}rinen(2010)Gutmann and
  Hyv{\"{a}}rinen]{2010Michael}
Gutmann, M. and Hyv{\"{a}}rinen, A.
\newblock Noise-contrastive estimation: {A} new estimation principle for
  unnormalized statistical models.
\newblock 2010.

\bibitem[Hadsell et~al.(2006)Hadsell, Chopra, and Lecun]{2006Hadsell}
Hadsell, R., Chopra, S., and Lecun, Y.
\newblock Dimensionality reduction by learning an invariant mapping.
\newblock In \emph{2006 IEEE Computer Society Conference on Computer Vision and
  Pattern Recognition (CVPR'06)}, 2006.

\bibitem[He et~al.(2016)He, Zhang, Ren, and Sun]{2016Kaiming}
He, K., Zhang, X., Ren, S., and Sun, J.
\newblock Deep residual learning for image recognition.
\newblock In \emph{2016 {IEEE} Conference on Computer Vision and Pattern
  Recognition, {CVPR} 2016}, 2016.

\bibitem[He et~al.(2020)He, Fan, Wu, Xie, and Girshick]{2020Kaiming}
He, K., Fan, H., Wu, Y., Xie, S., and Girshick, R.
\newblock Momentum contrast for unsupervised visual representation learning.
\newblock In \emph{2020 IEEE/CVF Conference on Computer Vision and Pattern
  Recognition (CVPR)}, 2020.

\bibitem[Hjelm et~al.(2018)Hjelm, Fedorov, Lavoie-Marchildon, Grewal, Bachman,
  Trischler, and Bengio]{hjelm2018learning}
Hjelm, R.~D., Fedorov, A., Lavoie-Marchildon, S., Grewal, K., Bachman, P.,
  Trischler, A., and Bengio, Y.
\newblock Learning deep representations by mutual information estimation and
  maximization.
\newblock \emph{arXiv preprint arXiv:1808.06670}, 2018.

\bibitem[Hénaff et~al.(2019)Hénaff, Srinivas, Fauw, Razavi, Doersch, Eslami,
  and Oord]{OJ2019Data}
Hénaff, O., Srinivas, A., Fauw, J.~D., Razavi, A., Doersch, C., Eslami, S.,
  and Oord, A.
\newblock Data-efficient image recognition with contrastive predictive coding.
\newblock 2019.

\bibitem[Jaderberg et~al.(2016)Jaderberg, Czarnecki, Osindero, Vinyals, Graves,
  and Kavukcuoglu]{2016DecoupledJaderberg}
Jaderberg, M., Czarnecki, W.~M., Osindero, S., Vinyals, O., Graves, A., and
  Kavukcuoglu, K.
\newblock Decoupled neural interfaces using synthetic gradients.
\newblock 2016.

\bibitem[Jia et~al.(2009)Jia, Wei, Socher, Li, Kai, and Li]{2009Feifei}
Jia, D., Wei, D., Socher, R., Li, L.~J., Kai, L., and Li, F.~F.
\newblock Imagenet: A large-scale hierarchical image database.
\newblock \emph{Proc of IEEE Computer Vision and Pattern Recognition}, 2009.

\bibitem[Krizhevsky et~al.(2009)Krizhevsky, Hinton,
  et~al.]{krizhevsky2009learning}
Krizhevsky, A., Hinton, G., et~al.
\newblock Learning multiple layers of features from tiny images.
\newblock 2009.

\bibitem[Krizhevsky et~al.(2012)Krizhevsky, Sutskever, and
  Hinton]{2012Krizhevsky}
Krizhevsky, A., Sutskever, I., and Hinton, G.
\newblock Imagenet classification with deep convolutional neural networks.
\newblock In \emph{NIPS}, 2012.

\bibitem[Li et~al.(2017)Li, Zhou, Fei, and Hang]{2017MetaLi}
Li, Z., Zhou, F., Fei, C., and Hang, L.
\newblock Meta-sgd: Learning to learn quickly for few-shot learning.
\newblock 2017.

\bibitem[Liu et~al.(2019)Liu, Davison, and Johns]{2019SelfLiu}
Liu, S., Davison, A.~J., and Johns, E.
\newblock Self-supervised generalisation with meta auxiliary learning.
\newblock 2019.

\bibitem[Ma et~al.(2018)Ma, Shen, Dick, Wu, Wang, Hengel, and
  Reid]{2018VisualMa}
Ma, C., Shen, C., Dick, A., Wu, Q., Wang, P., Hengel, A., and Reid, I.
\newblock Visual question answering with memory-augmented networks.
\newblock \emph{IEEE}, 2018.

\bibitem[Oord et~al.(2018)Oord, Li, and Vinyals]{2018RepresentationOord}
Oord, A. v.~d., Li, Y., and Vinyals, O.
\newblock Representation learning with contrastive predictive coding.
\newblock \emph{arXiv preprint arXiv:1807.03748}, 2018.

\bibitem[Robinson et~al.(2020)Robinson, Chuang, Sra, and Jegelka]{2020Hard}
Robinson, J., Chuang, C.~Y., Sra, S., and Jegelka, S.
\newblock Contrastive learning with hard negative samples.
\newblock 2020.

\bibitem[Schmidhuber(2014)]{2014Schmidhuber}
Schmidhuber, J.
\newblock Learning complex, extended sequences using the principle of history
  compression.
\newblock \emph{Neural Computation}, 4\penalty0 (2):\penalty0 234--242, 2014.

\bibitem[Schroff et~al.(2015)Schroff, Kalenichenko, and Philbin]{FaceNet2}
Schroff, F., Kalenichenko, D., and Philbin, J.
\newblock \emph{Facenet: A unified embedding for face recognition and
  clustering}.
\newblock 2015.

\bibitem[Snell et~al.(2017)Snell, Swersky, and Zemel]{2017Jake}
Snell, J., Swersky, K., and Zemel, R.~S.
\newblock Prototypical networks for few-shot learning.
\newblock In \emph{Advances in Neural Information Processing Systems 30: Annual
  Conference on Neural Information Processing Systems 2017, December 4-9, 2017,
  Long Beach, CA, {USA}}, 2017.

\bibitem[Sridharan \& Kakade(2008)Sridharan and Kakade]{2008Sridharan}
Sridharan, K. and Kakade, S.~M.
\newblock An information theoretic framework for multi-view learning.
\newblock \emph{Conference on Learning Theory}, 2008.

\bibitem[Strubell et~al.(2019)Strubell, Ganesh, and
  Mccallum]{2019EnergyStrubell}
Strubell, E., Ganesh, A., and Mccallum, A.
\newblock Energy and policy considerations for deep learning in nlp.
\newblock 2019.

\bibitem[Sun et~al.(2020)Sun, Cheng, Zhang, Zhang, Zheng, Wang, and
  Wei]{Circle2}
Sun, Y., Cheng, C., Zhang, Y., Zhang, C., Zheng, L., Wang, Z., and Wei, Y.
\newblock \emph{Circle loss: A unified perspective of pair similarity
  optimization}.
\newblock 2020.

\bibitem[Tian et~al.(2019)Tian, Krishnan, and Isola]{Tian2019Contrastive}
Tian, Y., Krishnan, D., and Isola, P.
\newblock Contrastive multiview coding.
\newblock \emph{arXiv preprint arXiv:1906.05849}, 2019.

\bibitem[Tian et~al.(2020)Tian, Sun, Poole, Krishnan, Schmid, and
  Isola]{2020WhatTian}
Tian, Y., Sun, C., Poole, B., Krishnan, D., Schmid, C., and Isola, P.
\newblock What makes for good views for contrastive learning.
\newblock 2020.

\bibitem[Tsai et~al.(2020)Tsai, Wu, Salakhutdinov, and Morency]{2020Tsai}
Tsai, Y., Wu, Y., Salakhutdinov, R., and Morency, L.~P.
\newblock Self-supervised learning from a multi-view perspective.
\newblock 2020.

\bibitem[Vaswani et~al.(2017)Vaswani, Shazeer, Parmar, Uszkoreit, Jones, Gomez,
  Kaiser, and Polosukhin]{2017Ashish}
Vaswani, A., Shazeer, N., Parmar, N., Uszkoreit, J., Jones, L., Gomez, A.~N.,
  Kaiser, L., and Polosukhin, I.
\newblock Attention is all you need.
\newblock 2017.

\bibitem[Verma et~al.(2021)Verma, Luong, Kawaguchi, Pham, and Le]{2021Vikas}
Verma, V., Luong, T., Kawaguchi, K., Pham, H., and Le, Q.~V.
\newblock Towards domain-agnostic contrastive learning.
\newblock In \emph{Proceedings of the 38th International Conference on Machine
  Learning, {ICML} 2021, 18-24 July 2021, Virtual Event}, Proceedings of
  Machine Learning Research. {PMLR}, 2021.

\bibitem[Vinyals et~al.(2016)Vinyals, Blundell, Lillicrap, Kavukcuoglu, and
  Wierstra]{2016Oriol}
Vinyals, O., Blundell, C., Lillicrap, T., Kavukcuoglu, K., and Wierstra, D.
\newblock Matching networks for one shot learning.
\newblock In \emph{Advances in Neural Information Processing Systems 29: Annual
  Conference on Neural Information Processing Systems 2016, December 5-10,
  2016, Barcelona, Spain}, 2016.

\bibitem[Wu et~al.(2018)Wu, Xiong, Yu, and Lin]{un2}
Wu, Z., Xiong, Y., Yu, S.~X., and Lin, D.
\newblock \emph{Unsupervised feature learning via non-parametric instance
  discrimination}.
\newblock 2018.

\bibitem[Xiao et~al.(2021)Xiao, Wang, Efros, and Darrell]{2021Tete}
Xiao, T., Wang, X., Efros, A.~A., and Darrell, T.
\newblock What should not be contrastive in contrastive learning.
\newblock In \emph{9th International Conference on Learning Representations,
  {ICLR} 2021, Virtual Event, Austria, May 3-7, 2021}. OpenReview.net, 2021.

\bibitem[Xu et~al.(2021)Xu, Zhou, Fu, Zhou, and Li]{2021AGDLJingjing}
Xu, J., Zhou, W., Fu, Z., Zhou, H., and Li, L.
\newblock A survey on green deep learning.
\newblock \emph{CoRR}, 2021.

\bibitem[Zbontar et~al.(2021)Zbontar, Li, Misra, Lecun, and Deny]{2021Barlow}
Zbontar, J., Li, J., Misra, I., Lecun, Y., and Deny, S.
\newblock Barlow twins: Self-supervised learning via redundancy reduction.
\newblock 2021.

\bibitem[Zhang et~al.(2017)Zhang, Isola, and Efros]{Splitp2}
Zhang, R., Isola, P., and Efros, A.~A.
\newblock Split-brain autoencoders: Unsupervised learning by cross-channel
  prediction.
\newblock 2017.

\bibitem[Zhang et~al.(2018)Zhang, Tang, and Jia]{2018FineYabin}
Zhang, Y., Tang, H., and Jia, K.
\newblock Fine-grained visual categorization using meta-learning optimization
  with sample selection of auxiliary data.
\newblock In \emph{Computer Vision - {ECCV} 2018 - 15th European Conference,
  Munich, Germany, September 8-14, 2018, Proceedings, Part {VIII}}, Lecture
  Notes in Computer Science, pp.\  241--256. Springer, 2018.

\end{thebibliography}
